\documentclass[lettersize,journal]{IEEEtran}
\usepackage{amsmath,amsfonts}
\usepackage{algorithmic}
\usepackage{algorithm}
\usepackage{array}
\usepackage[caption=false,font=normalsize,labelfont=sf,textfont=sf]{subfig}
\usepackage{textcomp}
\usepackage{stfloats}
\usepackage{url}
\usepackage{verbatim}
\usepackage{graphicx}
\usepackage{cite}
\usepackage[table]{xcolor}
\usepackage{mdframed}
\usepackage{ulem}
\usepackage{amssymb}
\usepackage{mathtools}
\usepackage{amsthm}
\usepackage{booktabs}
\usepackage{threeparttable}
\usepackage{wrapfig}
\usepackage{hyperref}
\usepackage{subfig} 
\usepackage{balance}
\usepackage{bm}

\DeclareMathOperator*{\argmax}{arg\,max}

\newcommand{\model}{TCTO}

\begin{document}

\title{Collaborative Multi-Agent Reinforcement Learning for Automated Feature Transformation with Graph-Driven Path Optimization}

\author{Xiaohan Huang, Dongjie Wang, Zhiyuan Ning, Ziyue Qiao, Qingqing Long, Haowei Zhu,\\Yi Du, Min Wu,~\IEEEmembership{Senior Member,~IEEE}, Yuanchun Zhou and Meng Xiao,~\IEEEmembership{Member,~IEEE} 
\IEEEcompsocitemizethanks{
\IEEEcompsocthanksitem Xiaohan Huang, Zhiyuan Ning and Qingqing Long are with the Computer Network Information Center, Chinese Academy of Sciences, and the University of the Chinese Academy of Sciences. Emails: xhhuang@cnic.cn, nzy@cnic.cn, qqlong@cnic.cn.
\IEEEcompsocthanksitem Yi Du, Yuanchun Zhou, and Meng Xiao are with the Computer Network Information Center, Chinese Academy of Sciences. Emails: duyi@cnic.cn, zyc@cnic.cn, shaow@cnic.cn.
\IEEEcompsocthanksitem Dongjie Wang is with the Department of Electrical Engineering and Computer Science at the University of Kansas. Email: wangdongjie@ku.edu. 
\IEEEcompsocthanksitem Ziyue Qiao is with the School of Computing and Information Technology, Great Bay University, Dongguan, China. Email: zyqiao@gbu.ecu.cn. 
\IEEEcompsocthanksitem Haowei Zhu is with the School of Software, Tsinghua University. E-mails: zhuhw23@mails.tsinghua.edu.cn.
\IEEEcompsocthanksitem Min Wu is with the Institute for Infocomm Research, Agency for Science, Technology, and Research. E-mails: wumin@i2r.a-star.edu.sg.
\IEEEcompsocthanksitem Corresponding author: Meng Xiao.
\IEEEcompsocthanksitem Our codes and data are publicly accessible via~\href{https://www.dropbox.com/scl/fi/ygvvmq7e7xto8q54b9ons/TCTO-Code.zip?rlkey=e2r67rhdd34ut7d85kwzjpiob&st=qve46gxm&dl=0}{Dropbox}.
}
}

\markboth{Journal of \LaTeX\ Class Files,~Vol.~14, No.~8, August~2021}%
{Shell \MakeLowercase{\textit{et al.}}: A Sample Article Using IEEEtran.cls for IEEE Journals}


\maketitle

\begin{abstract}
Feature transformation methods aim to find an optimal mathematical feature-feature crossing process that generates high-value features and improves the performance of downstream machine learning tasks.
Existing frameworks, though designed to mitigate manual costs, often treat feature transformations as isolated operations, ignoring dynamic dependencies between transformation steps.
To address the limitations, we propose \model, a collaborative multi-agent reinforcement learning framework that automates feature engineering through graph-driven path optimization.
The framework’s core innovation lies in an evolving interaction graph that models features as nodes and transformations as edges. 
Through graph pruning and backtracking, it dynamically eliminates low-impact edges, reduces redundant operation, and enhances exploration stability. 
This graph also provides full traceability to empower \model\ to reuse high-utility subgraphs from historical transformations.
To demonstrate the efficacy and adaptability of our approach, we conduct comprehensive experiments and case studies, which show superior performance across a range of datasets.
\end{abstract}




\section{Introduction}

\IEEEPARstart{C}{lassical} machine learning (ML) heavily relies on the structure of the model and the quality of the involving features~\cite{sambasivan2021everyone,strickland2022andrew,borisov2022deep,zha2023data}. 
This dependency makes designing effective features a crucial step before the learning process.
Traditionally, designing effective features required extensive manual intervention, where scientists applied mathematical transformations to raw data to create meaningful ones~\cite{conrad2022benchmarking,wang2025towards}. 
This process, illustrated in Figure~\ref{fig:motivation}, is known as \textit{feature transformation}~\cite{dong2018feature,nargesian2017learning,he2025fastft}. 
In contrast, models such as gradient-boosted trees (GBTs)~\cite{si2017gradient} and deep neural networks (DNNs)~\cite{bengio2013representation} can automatically capture non-linear feature interactions. 
However, these models typically require large-scale training data to generalize well~\cite{grinsztajn2022tree,shwartz2022tabular}. 
To enhance feature transformation and reduce reliance on manual engineering, recent studies have explored AI-driven \textit{automated feature transformation} (AFT)~\cite{zha2023data,cui2024tabular,wang2025datacentricaicomprehensivesurvey}, which aims to automatically generate, select, and refine features. 
These methods can be categorized into three approaches:
\begin{figure}[t]
\begin{center}
\centerline{\includegraphics[width=\columnwidth]{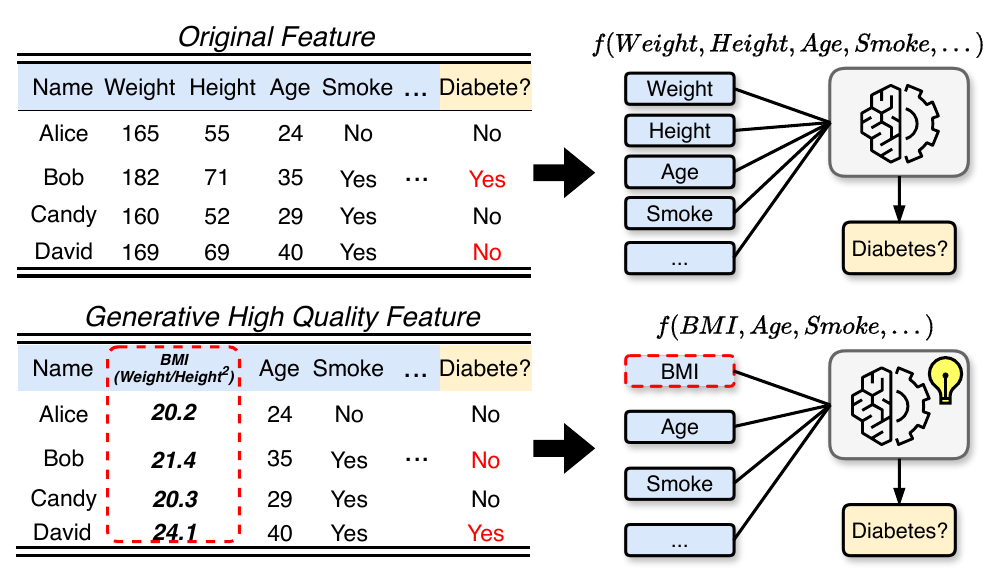}}
\caption{High-quality features contribute to the performance of machine learning models.}
\label{fig:motivation}
\end{center}
\vspace{-0.3in}
\end{figure}

\textit{1) expansion-reduction approaches}~\cite{kanter2015deep,khurana2016cognito,horn2019autofeat} randomly combine and generate features through mathematical transformations, then employ feature selection techniques to isolate high-quality features. 
Those approaches are highly stochastic, lack stability, and are not optimization-oriented. 
\textit{2) iterative feedback approaches}~\cite{tran2016genetic,li2023learning,liu2024interpretable} aim to refine the feature space with the transformation towards reinforcement learning ~\cite{kdd2022,xiao2022traceable,xiao2024traceable,ying2023self}. 
Although those methods can optimize their strategies during the exploration, they discard the valuable experiences from historical sub-transformations and treat feature transformations as isolated operations. 
One pioneer study~\cite{khurana2018feature} maintains a transformation graph to store the exploration experiences.
However, it focuses on the dataset-level transformation and lacks a backtracking mechanism, which constrains the flexibility and stability of the transformation process.
\textit{3) AutoML approaches}~\cite{zhu2022difer, zhang2023openfe} partially adjust aforementioned issues by learning the pattern of the collected historical transformation records~\cite{wang2024reinforcement} thus reach a so-called global view of the action space~\cite{ying2024unsupervised}.
Nevertheless, a clear disadvantage of these methods is that they initially rely on the quality of collected transformations to construct a search space that closely mirrors real-world feature representations. 

Building on these discussions, we revisit the iterative feedback approaches and introduce Flexible \textbf{T}ransformation-\textbf{C}entric \textbf{T}abular Data \textbf{O}ptimization Framework (\textbf{\model}), a novel automated feature transformation methodology employing collaborative multi-agent reinforcement learning (MARL)~\cite{busoniu2008comprehensive, panait2005cooperative}.
Our framework is structured around an evolving feature transformation graph maintained throughout the MARL process.
This graph serves as a comprehensive \textit{roadmap}, where each node and its path back to the root node represents a transformation pathway applied to the dataset's initial features.
Our optimization procedure comprises four steps:
(1) clustering each node on the roadmap with mathematical and spectral characteristics;
(2) state representation for each cluster;
(3) cluster-level transformation decision generation based on multi-agent collaboration;
(4) evaluation and reward estimation for the generated outcomes.
Iteratively, \model\ executes these steps while leveraging the traceability of the roadmap for precise node-wise and step-wise pruning.
This allows for redundant transformation reduction,  strategic rollbacks, and transformation pathway optimization.

\begin{figure}[t]
\begin{center}
\centerline{\includegraphics[width=\columnwidth]{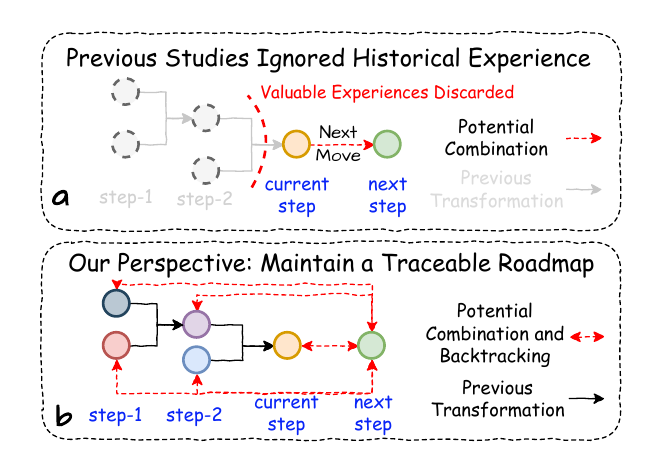}}
\caption{The technical contributions summarization.}
\label{fig:technical}
\end{center}
\vskip -0.4in
\end{figure}

\textbf{Our contribution and perspective:} As shown in Figure~\ref{fig:technical}, our method delivers three key benefits for feature transformation that address critical limitations in previous approaches:

(1) \textit{Enhanced Transformation Agility:} \model\ is designed to capture and dynamically apply transformations across various stages of the feature transformation process rather than being restricted to transformations derived from the latest feature set. Unlike previous studies that ignored historical features (Figure~\ref{fig:technical}a), our approach maintains a traceable roadmap of all transformation steps. This enables a more flexible and robust handling of features through potential combinations and backtracking when needed.

(2) \textit{Historical Insights Utilization:} We leverage deep learning techniques to extract and model latent correlations and mathematical characteristics from past transformation efforts. Prior methods often discarded valuable experiences (Figure~\ref{fig:technical}a), our system actively preserves and utilizes these information. The historical insight allows the RL agents to maintain a global view of the transformation pathway (Figure~\ref{fig:technical}b) and make decisions based on comprehensive learned experience rather than just the current state.

(3) \textit{Robust Backtracking Mechanism:} Our approach incorporates a sophisticated backtracking system that utilizes historical transformation records for traceability. As illustrated in Fig~\ref{fig:technical}b, this includes tracking previous transformations and enabling transitions between different steps in the transformation sequence. This ensures that the transformation process can revert or alter its course to avoid inefficient or suboptimal trajectories, thus optimizing the overall feature engineering pathway. The system can intelligently combine previous transformations or backtrack when necessary, overcoming the sequence progression limitation of conventional methods.

Through rigorous experimental validation, we demonstrate that \model\ can enhance the flexibility of the optimization process and deliver more resilient and effective results compared to conventional iterative optimization frameworks.

\section{preliminary}
\subsection{Important Definitions}

\noindent\textbf{Input Dataset.} 
Formally, a  dataset can be defined as $\mathcal{D} = [\mathcal{F}, Y]$, where $\mathcal{F}=\{f_1,\dots, f_n\}$ represents $n$ features and $Y$ stands for the labels. 
Each row of $\mathcal{D}$ represents a single observation or data point, while each column corresponds to a specific attribute or feature of the observation. 

\noindent\textbf{Operation Set.} 
Mathematical operations can be applied to the existing features, generating new and informative-derived features. 
We define this collection of operations as the operation set, represented by the symbol $\mathcal{O}$. 
The operations within this set can be categorized into two main types according to their computational properties: unary and binary operations.
The details of the operation set are listed as follows.
The token $x$ is a scalar, which implies each element in vector $X$.
\begin{itemize}
    \item Elementary mathematical operation
    \begin{itemize}
        \item Unary: $x^2 , x^3, \sqrt{x}, \sin{x}, \cos{x}, \log_e(x), e^x$
        \item Binary: $+,-,\times,\div$
    \end{itemize}
    \item Functional mathematical operation
    \begin{itemize}
        \item tanh: $x' = \frac{e^x - e^{-x}}{e^x + e^{-x}}$
        \item sigmoid: $x' = \frac{1}{1 + e^{-x}}$
        \item reciprocal: $x' = \frac{1}{x}$
        \item stand\_scaler: $X' = \frac{X - \mu}{\sigma}$\\ 
        Note: $\mu$ and $\sigma$ is the mean and standard deviation of $X$, respectively.
        \item minmax\_scaler: $X' = \frac{X - X_{min}}{X_{max}-X_{min}}$ \\
        Note: $X_{min}$ and $X_{max}$ mean the minus and max element of $X$, respectively.
        \item quantile\_transform: $X'=Quantile(X)$ \\
        Note: Quantile transforms features to follow a uniform distribution.
    \end{itemize}
\end{itemize}
\begin{figure}[t]
\begin{center}
\centerline{\includegraphics[width=\columnwidth]{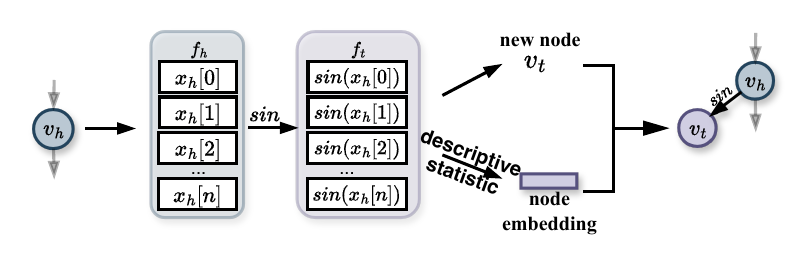}}
\caption{An example of feature transformation roadmap update: the feature $f_h$ conducts $sin$ operation generating the feature $f_t$. 
  The embedding of node $v_t$ can be derived from the statistic description of generated feature $f_t$.}
\label{dem01}
\end{center}
\vskip -0.2in
\end{figure}
\begin{figure*}[!t]
\begin{center}
\centerline{\includegraphics[width=\linewidth]{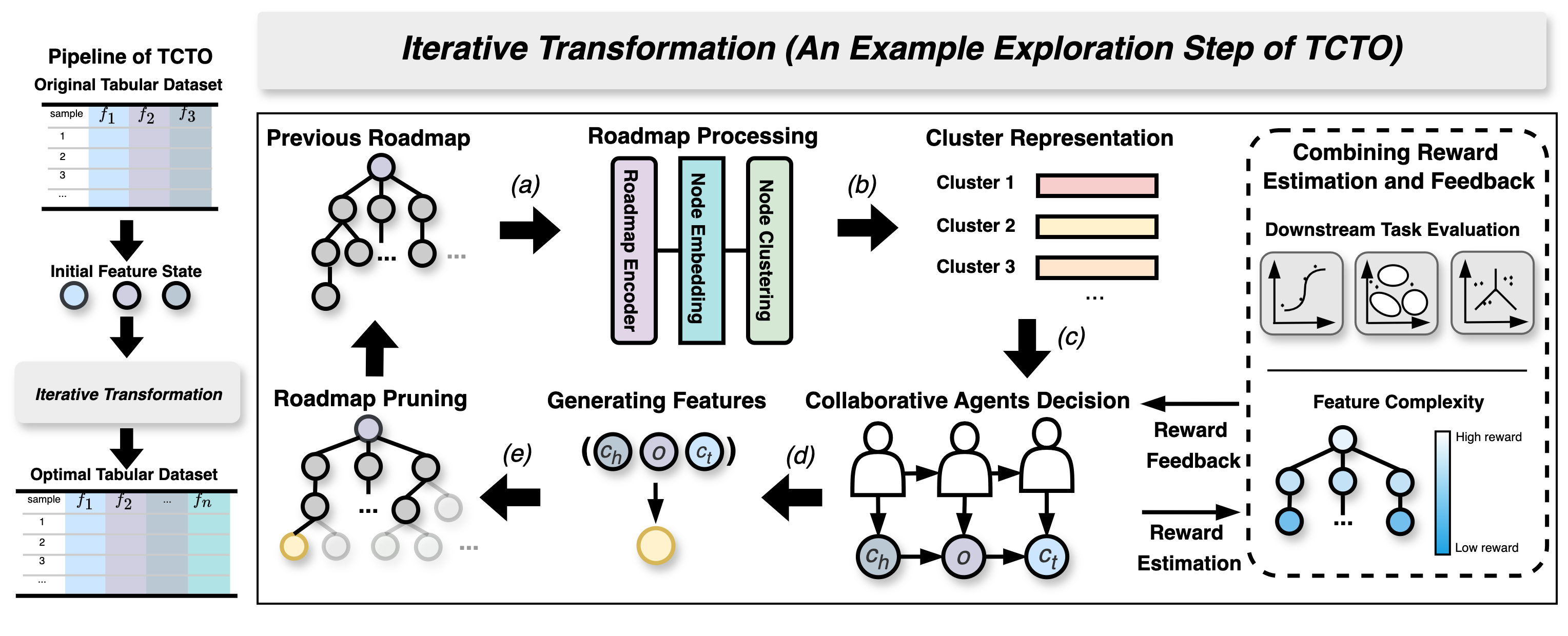}}
\caption{An overview of our framework: (a) cluster and represent the nodes on roadmap; (b) represent the node clusters; (c) reinforce multi-agent feature transformation decision generation; (d)generate high-quality features; (e) prune the roadmap.}
\label{fig:main}
\end{center}
\vskip -0.3in
\end{figure*}

\noindent\textbf{Feature Transformation Roadmap.}
A feature transformation roadmap $\mathcal{G}$ is an evolving directed graph and could uniquely represent the global optimization process.
Figure~\ref{dem01} shows an example of the new generation of edges and nodes. 
This roadmap, denoted as $\mathcal{G}  = \{V, E, \mathcal{A}\}$, consists of multiple tree structures where the number of trees equals the number of features in the original dataset. 
$V = \{v_i\}_{i=1}^m$ and $E = \{e_i\}_{i=1}^n$ represent the set of feature state nodes\footnote{Note that in the formulas, \( v \) also represents the embedding of node \( v \) for the sake of simplification. } and transformation edges, respectively. $\mathcal{A}$ is the adjacency matrix.
Each pair of nodes, connected by a directed edge, represents a new feature state $v_t$ generated from a previous state $v_h$ after undergoing the transformation represented by the type of edge $e$. 
The embedding of each node will be obtained via the descriptive statistics information (e.g., the standard deviation, minimum, maximum, and the first, second, and third quartile) of the generated features.
We can apply the roadmap to generate a new dataset $\mathcal{D}'$ with a given dataset, defined as $\mathcal{D}' = \mathcal{G}(\mathcal{D})$.
\subsection{Feature Transformation Problem}
As the example illustrated in Figure~\ref{fig:motivation}, given a downstream target ML model $\mathcal{M}$ (e.g., classification model, regression model, etc.) and a dataset $\mathcal{D} = [\mathcal{F}, Y]$, the objective is to find an optimal feature transformation roadmap $\mathcal{G}^*$ that can optimize the dataset through mathematical operation in $\mathcal{O}$.
Formally, the objective function can be defined as:
\begin{equation}
    \label{objective}
    \mathcal{G}^{*} = \argmax_{\mathcal{G}} \mathcal{V}( \mathcal{M}(\mathcal{G}(\mathcal{F})),Y),
\end{equation}
where $\mathcal{V}$ denotes the evaluation metrics according to the target downstream ML model $\mathcal{M}$.

\section{Proposed method}
Figure~\ref{fig:main} provides an overview of the \model\ framework, which outlines the iterative process guided by reinforcement learning (RL). 
We begin with a comprehensive introduction to clustering on the transformation roadmap and representing the roadmap state for each agent.
Next, we delve into the multi-agent learning system and the process of generating group-wise operations.
Finally, we introduce reward estimation and policy optimization.

\subsection{Group-wise Transformation}
\label{method_cluster}
Group-wise transformation plays a pivotal role in enhancing the efficiency and effectiveness of the transformation process. 
The nodes on the roadmap are first clustered and then represented using network encoding. The details are as follows.

\noindent\textbf{Roadmap Clustering.} Group-wise operations are motivated by the observation that mathematical interactions between distinct feature groups often generate highly informative features~\cite{kdd2022}. 
By clustering features with inherent similarities, their combined transformation potential can be leveraged to produce richer and more discriminative feature representations.
Inspired by~\cite{von2007tutorial}, a similarity matrix $\tilde{\mathcal{A}}$ is computed based on the cosine similarity between the embedding vectors of the nodes, given by:
\begin{equation}
\tilde{\mathcal{A}}[i, j] = \frac{v_i \cdot v_j}{\|v_i\| \|v_j\|}
\end{equation}
Additionally, an enhanced Laplacian matrix $\mathcal{S}$ is defined to capture both structural and mathematical information from the nodes, formulated as follows:
\begin{equation}
\mathcal{S} = \mathcal{D} - (\mathcal{A} +  \tilde{\mathcal{A}})
\end{equation}
Here, $\mathcal{D}$ represents the degree matrix, with diagonal elements $\mathcal{D}[i, i]$ equal to the sum of the elements in the $i$-th row of $\mathcal{A} + \tilde{\mathcal{A}}$.
The clustering module uses hierarchical clustering based on the eigenvalues and eigenvectors of $\mathcal{S}$ to identify the optimal roadmap partition into clusters.
The clustering module treats each eigenvector corresponding to node $v_{i}$ as an initial singleton cluster and iteratively merges pairs of shortest clusters to progressively form larger clusters. 
This process continues until the cluster number reaches a specified setting, denoted as $k$.
The set of clusters is denoted as $C = \{c_i\}_{i=1}^{k}$.

\noindent\textbf{Roadmap-based State Representation.}
The transformation roadmap serves as a repository of intermediate transformation records, capturing the historical evolution of feature transformations. To extract the latent correlations within these historical records and effectively represent the state of each feature cluster, our framework integrates a Relational Graph Convolutional Network (RGCN)~\cite{schlichtkrull2018modeling}. 
Specifically, a dual-layer RGCN framework was conducted to disseminate information across nodes, utilizing various relationship types to accurately represent the state of each node and transformation operation, described as:
\begin{equation}
    v_i^{(l+1)} = \phi \left( \sum_{r=1}^{R} \sum_{j \in N^{r}_{i}} \frac{1}{c_{i,r}} W_r^{(l)} v_j^{(l)} \right)
\end{equation}
where $v_i^{(l)}$ and $v_i^{(l+1)}$ represents the embedding of the $i$-th node in the roadmap at RGCN layer-$l$ and layer-$(l+1)$, respectively.
$N^{r}_{i}$ denotes the set of neighboring nodes of $v_i$ with operation type $r$, and the degree normalization factor $c_{i,r}$ scales the influence of neighboring nodes. 
$r$ represents the relationships between nodes, which correspond to different operation types.
The resulting sum is then passed through an activation function $\phi$ to produce the final representation of the node $v_i$.
Based on the aggregated node representation, the state representation of the cluster $c_i$ can be obtained by $Rep(c_i) = \frac{1}{|c_i|}\sum_{v\in c_i} v$, 
where $|c_i|$ denotes the number of nodes in cluster $c_i$.

\subsection{Collaborative Multi-agent Architecture}

Three agents are employed that collaboratively construct mathematical transformations. These agents operate sequentially to select the optimal head cluster, mathematical operation, and operand cluster, respectively. 
Specifically, Figure~\ref{decision} shows an example of the collaborative agents' decision-making process. 
A series of collaborative agents operate sequentially, each performing a specific task, with the output of one agent serving as the input for the next in a step-by-step decision-making process.
The details of each agent are as follows:
\begin{figure}[t]
\begin{center}
\centerline{\includegraphics[width=\linewidth]{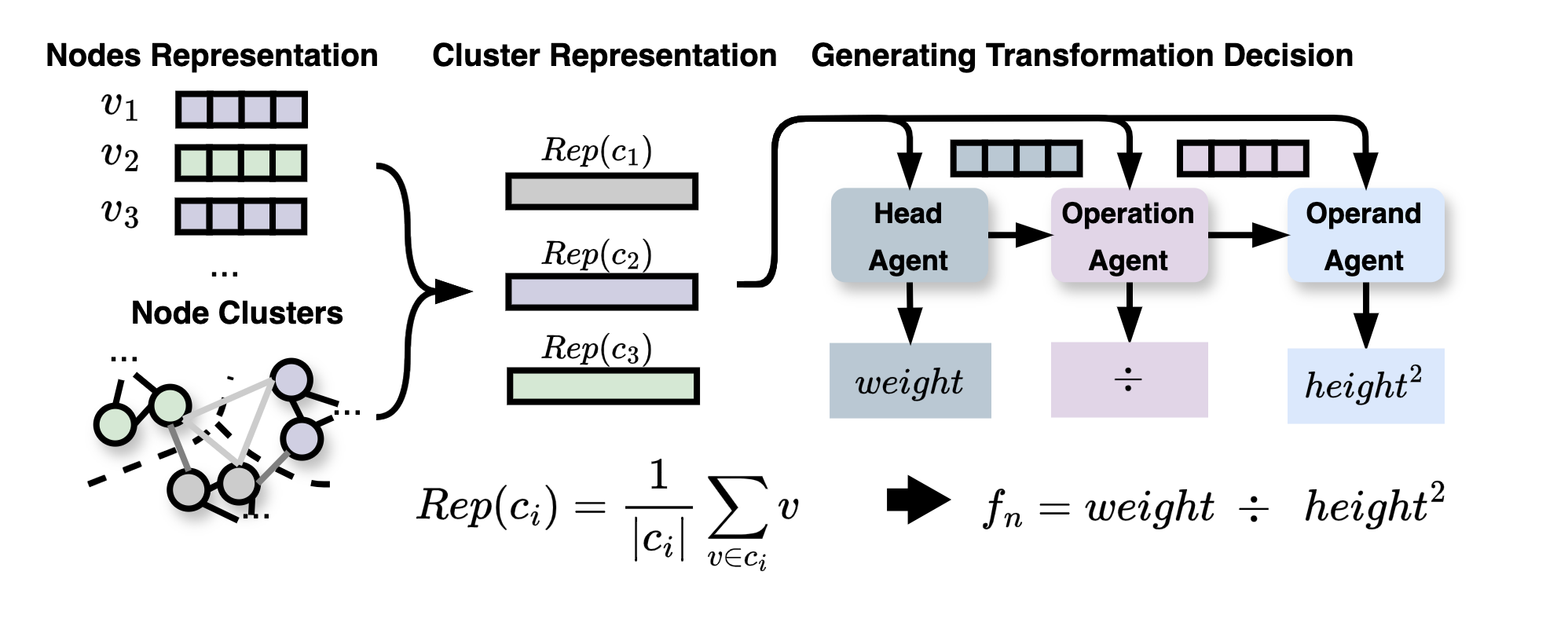}}
\caption{The reinforcement learning decision process. 
Three agents collaborate to generate a binary transformation.}
\label{decision}
\end{center}
\vskip -0.3in
\end{figure}

\noindent\textbf{Head Cluster Agent.}
After each node on the roadmap has been clustered, denoted as $\mathcal{C}$,
the first agent aims to select the head cluster to be transformed according to the current state of each cluster. 
Specifically, the $i$-th cluster state is given as $Rep(c_i)$, and the overall state can be represented as $Rep(V)$.
With the head policy network $\pi_h(\cdot)$, the score of select $c_i$ as the action can be estimated by: $s_i^h = \pi_h(Rep(c_i)\oplus Rep(V))$. 
The token $c_h$ denotes the selected cluster with the highest score and $\oplus$ indicates vector concatenation.

\noindent\textbf{Operation Agent.}
The operation agent aims to select the mathematical operation to be performed according to the overall roadmap and selected head cluster. 
The policy network in the operation agent takes $Rep(c_h)$ and the global roadmap state as input, then chooses an optimal operation from the operation set $\mathcal{O}$: $o = \pi_o(Rep(c_h)\oplus Rep(V))$.

\noindent\textbf{Operand Cluster Agent.}
If the operation agent selects a binary operation, the operand cluster agent will choose a tail cluster to perform the transformation. 
The policy network $\pi_t(\cdot)$ will take the state of the selected head cluster, the operation, the global roadmap state, and the $i$-th candidate tail cluster as input, given as $s_i^t = \pi_t(Rep(c_h)\oplus Rep(V)\oplus Rep(o)\oplus Rep(c_i))$, where $Rep(o)$ is a one-hot embedding for each operation. 
The token $c_t$ denotes the selected tail cluster with the highest score. 

\noindent\textbf{Group-wise Feature-Crossing.} 
These stages above are referred to as one exploration step. 
Depending on the selected head cluster $c_h$, operation $o$, and optional operand cluster $c_t$, \model\ will cross each feature and then update the transformation roadmap (as shown in Figure~\ref{dem01} and Figure~\ref{decision}).

\subsection{Dual Reward Estimation for Optimizing Agents}
\model\ evaluates the generated features via the performance of downstream tasks, meanwhile factoring the complexity of the generated features into the reward function. 
This dual focus on performance and complexity ensures that the model aims at effectiveness while avoiding overly complex solutions that could hinder practical traceability.

\noindent\textbf{Duel-Facet Reward Estimation.} 
The policy's overarching goal is to iteratively improve the actions taken by these agents to maximize the cumulative rewards derived from the performance of downstream tasks and the complexity of transformations in the roadmap.
The performance of downstream tasks and the complexity of the transformation roadmap are considered as rewards to optimize the reinforcement learning framework, denoted as $\mathcal{R}_p$ and $\mathcal{R}_c$, respectively. 

\textit{(1) Performance of Downstream Tasks: } As the objective in Equation~\ref{objective}, $\mathcal{R}_p$ is calculated as follows:
\begin{equation}
    \mathcal{R}_p = \mathcal{V}(\mathcal{M}(\mathcal{F}_{t+1}), Y) - \mathcal{V}(\mathcal{M}(\mathcal{F}_{t}), Y),
\end{equation}
where $\mathcal{F}_t$ indicates the feature set at the $t$-th step. 

\textit{(2) Complexity of the Transformation: } 
The feature complexity reward $\mathcal{R}_c$ is defined as follows:
\begin{equation}
    \mathcal{R}_c = \frac{1}{n} \sum_{j=1}^n \frac{1}{e^{h(v_j)}},
\end{equation}
where $h(v_j)$ represents the number of levels from the root node to node $v_j$ on $\mathcal{G}$.
The total reward $\mathcal{R}$ is defined as follows: $\mathcal{R} = \mathcal{R}_p + \mathcal{R}_c$. 
In each step, the framework assigns the reward equally to each agent that has action. 

\noindent\textbf{Optimization of the Pipeline.} 
\noindent\textit{1) Policy Optimization:} The learning process for each agent is driven by a reward mechanism that quantifies the effectiveness and efficiency of the transformations applied to the roadmap. Specifically, the optimization policy is framed within a value-based reinforcement learning approach, leveraging a dual network setup architecture: a prediction network and a target network. 
The prediction network generates action-value (Q-value) predictions that guide the agents' decision-making processes at each step. 
It evaluates the potential reward for each possible action given the current state, facilitating the selection of actions that are anticipated to yield the highest rewards.
The target network serves as a stable benchmark for the prediction network and helps to calculate the expected future rewards. 
Decoupling the Q-value estimation from the target values is crucial to reducing overestimations and ensuring stable learning. 
\noindent\textit{2) Loss Function:}
The loss function used for training the prediction network is defined as follows:
\begin{equation}
    \mathcal{L} = \left((\mathcal{Q}^\pi_p(s_t, a_t) - \left(\mathcal{R}_t + \gamma \cdot \max_{a_{t+1}} \mathcal{Q}^\pi_t(s_{t+1}, a_{t+1}) \right)\right)^2,
\end{equation}
where prediction network $\mathcal{Q}^\pi_p(s_t, a_t)$ is the Q-value for the current state-action pair from the policy network $\pi(\cdot)$. $\mathcal{R}_t$ is the immediate reward received after taking action $a_t$ in state $s_t$, and $\gamma$ is the discount factor.
The item $\max_{a_{t+1}} \mathcal{Q}^\pi_t(s_{t+1}, a_{t+1})$ is the maximum predicted Q value for the next state-action pair as estimated by the target network. 
The parameters of the prediction network are updated through gradient descent to minimize loss, thereby aligning the predicted Q values with the observed rewards plus the discounted future rewards. To maintain the stability of the learning process, the parameters of the target network are periodically updated by copying them from the prediction network.

\subsection{Prune Roadmap for Effective and Stable Transformation}

\begin{figure}[t]
\begin{center}
\centerline{\includegraphics[width=\columnwidth]{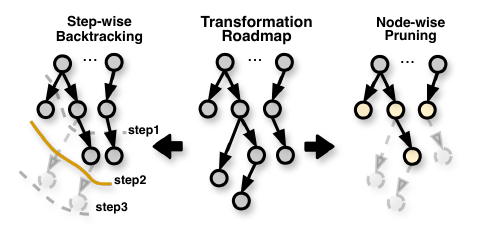}}
\caption{The two transformation roadmap pruning strategies.}
\label{prune}
\end{center}
\vskip -0.2in
\end{figure}

As illustrated in Figure~\ref{prune}, we employ two pruning strategies to ensure stability during the feature transformation process. 
\model\ adaptively adopts these strategies to reduce the potential explosion in  complexity, ensure the system remains efficient and manageable, and enhance stability. 

\noindent\textbf{Node-wise Pruning Strategy.} The node-wise pruning strategy will entail the identification of $K$ nodes that show the greatest relevance to labels.
This strategy computes the mutual information, defined as the relevance between each node's corresponding features and labels, as follows:
\begin{equation}
    \mathcal{I}(v,Y) = \sum_{f_i \in v} \sum_{y_i \in Y} p(f_i,y_i) \log \frac{p(f_i,y_i)}{p(f_i)p(y_i)}
\end{equation}
where $f_i$ denotes the element values of node $v$ and $y_i$ is its correlated label. 
The item $\mathcal{I}(v,Y)$ denotes the mutual information based score.
The token $p(f)$ represents the marginal probability distribution, while $p(f,y)$ represents the joint probability distribution. 
Finally, the framework will select top-$K$ nodes by the score.
The node-wise pruning strategy removes low-correlation nodes, preserving information as much as possible and ensuring exploration diversity.

\noindent\textbf{Step-wise Backtracking Strategy.} 
The step-wise backtracking strategy involves rolling back to the previous optimal transformation roadmap to prevent deviating onto suboptimal paths. 
This strategy ensures that the exploration process remains on the correct trajectory by revisiting and affirming the most effective roadmap configurations.

\noindent\textbf{Adaptively Pruning the Roadmap.} 
Pruning is applied when the number of nodes exceeds a predefined threshold to manage the complexity of the roadmap. During the early stages, when agents are less familiar with the dataset, the node-wise pruning strategy preserves diversity while reducing computational complexity. This approach ensures that the agents can explore without becoming overwhelmed by an excessively large feature space.
As the agents become more proficient and better understand the fundamental policy, \model\ transforms to a step-wise backtracking strategy. This strategy enhances the stability of exploration by refining the agent’s path and focusing on promising areas of the feature space.
By combining both pruning strategies, we strike a balance between enabling the agent to explore a sufficiently large search space early on, and ensuring stable, focused exploration in later stages of training. 
Specifically, \model\ applies node-wise pruning during the first 30\% of the exploration period and adopt step-wise backtracking for the remaining 70\% of the training process.

\subsection{The time complexity analysis of TCTO}

The time complexity of the proposed method can be analyzed across different processes. 
The clustering process involves eigen decomposition of a matrix, where the computational complexity is typically $\mathcal{O}(m^3)$, with $m$ representing the feature number. 
However, empirical results indicate that this step runs efficiently in practice. 
The decision process, which consists of neural network forward inference, depends on factors such as latent dimension and network architecture. 
The roadmap updating process prevents redundant node additions by comparing newly generated nodes with existing ones. 
Given a generative feature number $k$, the time complexity of this step is $\mathcal{O}(mk)$. 
The roadmap pruning process follows two strategies: node-wise pruning, which selects the most informative nodes with a complexity of $\mathcal{O}(nm)$, where $n$ is the sample number, and step-wise pruning, which operates in constant time $\mathcal{O}(1)$. 
In the downstream task process, taking the random forest algorithm as an example, tree construction and training require $\mathcal{O}(Tmn\log n)$, while model inference operates with $\mathcal{O}(T\log n)$ complexity, where $T$ denotes the number of constructed trees.

The space complexity of \model\ remains constant regardless of dataset size, as its reinforcement learning structure is fixed and independent of data scale. 
The head cluster agent's space complexity depends on the dimensions of the RGCN's hidden and output layers, as it encodes the roadmap in the initial step. 
Similarly, the operation agent's parameter scale is determined by the RGCN output dimension, as it makes decisions based on state information derived from roadmap embeddings. 
The operand cluster agent requires additional space due to an extra embedding layer for encoding mathematical operations within the value network. 
Furthermore, the model employs a dual value-network structure within the deep Q-learning framework, leading to a total parameter count that is twice the sum of the three cascading agents.



\section{experiment setting}

\subsection{Dataset and Evaluation Metrics}
\label{datasource}
Table~\ref{main_table} provides a summary of datasets, detailing sample sizes, feature dimensions, and task categories.
Our experimental analysis incorporated 16 classification datasets and 9 regression datasets.
The datasets were obtained from publicly accessible repositories, including Kaggle~\cite{kaggle}, LibSVM~\cite{libsvm}, OpenML~\cite{openml}, AutoML~\cite{automlchallenges} and the UCI Machine Learning Repository~\cite{uci}.
For evaluation, we utilized the F1-score for classification tasks and the 1-Relative Absolute Error (1-RAE) for regression tasks. In both cases, a higher value of the evaluation metric indicates that the generated features are more discriminative and effective.

\begin{table*}[h]
\centering
\caption{Overall performance comparison. `C' for binary classification and `R' for regression. The best results are highlighted in \textbf{bold}. The second-best results are highlighted in \underline{underline}. (\textbf{Higher values indicate better performance.}) \#Samp and \#Feat denote the number of samples and features.}
\label{main_table}
\resizebox{\linewidth}{!}{
\begin{tabular}{ccccccccccccccc}
\bottomrule
Dataset        & C/R & \#Samp. & \#Feat. & RDG  & ERG  & AFAT   & NFS   & TTG  & GRFG  & DIFER  & FETCH & OpenFE & FastFT &\model   \\  \midrule
Housing Boston    & R   & 506     & 13 & 0.404 & 0.409  & 0.416 & 0.425 & 0.396 & 0.465 & 0.381 & 0.440& 0.387 & \underline{0.491} & $\textbf{0.495}^{\pm 0.015}$\\ \hline
Airfoil           & R   & 1503    & 5  & 0.519 & 0.519  & 0.521 & 0.519 & 0.500 & 0.538 & 0.558 &0.601&\underline{0.605} & 0.585 &$\textbf{0.622}^{\pm 0.011}$ \\ \hline
Openml\_618       & R   & 1000    & 50 & 0.472 & 0.561 & 0.472 & 0.473 & 0.467 & \underline{0.589} & 0.408 &0.565&0.393 & 0.544 &$\textbf{0.600}^{\pm 0.005}$ \\ \hline
Openml\_589       & R   & 1000    & 25 & 0.509 & 0.610  & 0.508 & 0.505 & 0.503 & 0.599 & 0.463 &0.575&0.539 & \textbf{0.606} &$\textbf{0.606}^{\pm 0.003}$\\\hline
Openml\_616       & R   & 500     & 50 & 0.070 & 0.193 & 0.149 & 0.167 &  0.156 & \underline{0.467} & 0.076 &0.188 &0.100 & 0.452 &$\textbf{0.499}^{\pm 0.052}$\\ \hline
Openml\_607       & R   & 1000    & 50 & 0.521 & 0.555 & 0.516 & 0.519 &  0.522 & \underline{0.640} & 0.476 &0.571&0.430 & 0.579 &$\textbf{0.670}^{\pm 0.008}$\\ \hline
Openml\_620       & R   & 1000    & 25 &  0.511 & 0.546 & 0.527 & 0.513 & 0.512 &  \underline{0.626} & 0.442 &0.538 &0.489 & 0.620 &$\textbf{0.629}^{\pm 0.001}$\\ \hline
Openml\_637       & R   & 500     & 50 & 0.136 & 0.152 & 0.176 & 0.152 & 0.144 & 0.289 & 0.072  & 0.170& 0.055 & \underline{0.339} &$\textbf{0.355}^{\pm 0.022}$\\ \hline
Openml\_586       & R   & 1000    & 25 & 0.568 & 0.624 & 0.543 & 0.544 & 0.544 &  \underline{0.650} & 0.482 &0.611 &0.512 & 0.563 &$\textbf{0.689}^{\pm 0.004}$\\ \hline
Higgs Boson       & C  & 50000   & 28 & 0.695 & 0.702 & 0.697 & 0.691 & 0.699 & \textbf{0.709} & 0.669 & 0.697 & 0.702 & 0.692 & $\textbf{0.709}^{\pm 0.001}$ \\ \hline
Amazon Employee   & C   & 32769   & 9  & 0.932 & 0.934 & 0.930 & 0.932 & 0.933 & \underline{0.935} & 0.929 & 0.928& 0.931 & \underline{0.935} & $\textbf{0.936}^{\pm 0.001}$ \\ \hline
PimaIndian        & C   & 768     & 8 & 0.760 & 0.761 & 0.765 & 0.749 & 0.745 & 0.823 & 0.760 &0.774&0.744 & \underline{0.835} & $\textbf{0.850}^{\pm 0.007}$ \\ \hline
SpectF            & C   & 267     & 44 & 0.760 & 0.757 & 0.760 & 0.792 & 0.760 & 0.907 & 0.766 & 0.760&0.760 & \underline{0.927}  &$\textbf{0.950}^{\pm 0.012}$\\ \hline
SVMGuide3         & C & 1243 & 21 & 0.787 & 0.826 & 0.795 & 0.792 & 0.798 &  \underline{0.836} & 0.773 &0.772&0.810 & 0.835 & $\textbf{0.841}^{\pm 0.012}$\\\hline
German Credit     & C   & 1001    & 24   & 0.680 & 0.740 & 0.683 & 0.687 & 0.645 & 0.745 & 0.656 & 0.591& 0.706 & \underline{0.749} &$\textbf{0.768}^{\pm 0.008}$\\ \hline
Credit Default    & C   & 30000   & 25     & 0.805 & 0.803  & 0.804 & 0.801 & 0.798 & \underline{0.807} & 0.796 &0.747&0.802 & 0.800 & $\textbf{0.808}^{\pm 0.001}$\\ \hline
Messidor\_features& C   & 1150    & 19   & 0.624 & 0.669 & 0.665 & 0.638 & 0.655 & 0.718 & 0.660 & 0.730&0.702 & \underline{0.731} &$\textbf{0.742}^{\pm 0.003}$\\ \hline
Wine Quality Red  & C   & 999     & 12 & 0.466 & 0.461 & 0.480 & 0.462 & 0.467 & \underline{0.568} & 0.476 &0.510 &0.536 & \underline{0.568} &$\textbf{0.579}^{\pm 0.003}$\\ \hline
Wine Quality White& C   & 4900    & 12 & 0.524 & 0.510 & 0.516 & 0.525 & 0.531 & 0.543 & 0.507 & 0.507& 0.502 & \underline{0.551} & $\textbf{0.559}^{\pm 0.003}$\\ \hline
SpamBase          & C   & 4601    & 57 & 0.906 & 0.917 & 0.912 & 0.925 & 0.919 & 0.928 & 0.912 & 0.920&0.919 & \underline{0.929} &$\textbf{0.931}^{\pm 0.002}$ \\ \hline
AP-omentum-ovary  & C   & 275    & 10936  & 0.832 & 0.814  & 0.830 & 0.832 & 0.758 & \underline{0.868} & 0.833 &0.865&0.813 & 0.813 & $\textbf{0.888}^{\pm 0.002}$ \\ \hline
Lymphography      & C   & 148     & 18  & 0.108 & 0.144 & 0.150 & 0.152 & 0.148 & 0.342 &  0.150 & 0.158&\underline{0.379} & 0.353 & $\textbf{0.389}^{\pm 0.016}$\\ \hline
Ionosphere        & C   & 351     & 34 & 0.912 & 0.921& 0.928 & 0.913 & 0.902 & \textbf{0.971} & 0.905  & 0.942&0.899 & \textbf{0.971} &$\textbf{0.971}^{\pm 0.001}$\\ \hline
ALBERT$^{*}$           &  C   & 425240  & 78 & 0.678 & 0.619  & - & 0.680 & \underline{0.681} & - & - & - & 0.679 & 0.680 & $\textbf{0.681}^{\pm 0.004}$ \\ \hline
Newsgroups$^{*}$      & C   & 13142  &  61188 & 0.556 & 0.545  & 0.544 & 0.553 & 0.546 & - & - & - & 0.544 & \underline{0.580} & $\textbf{0.586}^{\pm 0.010}$ \\ 
\toprule
\end{tabular}}
\vspace{-0.1cm}
\begin{tablenotes}\footnotesize
    \item { We report F1-score for classification tasks and 1-RAE for regression tasks.}
    \item { The standard deviation is computed based on the results of 5 independent runs.}
    \item { * ALBERT and Newsgroups are large datasets, `-' indicates that the method ran out of memory or took too long.}
\end{tablenotes}
\vspace{-0.5cm}
\end{table*}

\subsection{Baseline Methods}
\label{baseline}
We conducted a comparative evaluation of {\model} against ten other automated feature transformation methods:
(1) \textbf{RDG} randomly selects an operation and applies it to various features to generate new transformed features.
(2) \textbf{ERG} conducts operations on all features simultaneously and selects the most discriminative ones as the generated features.
(3) \textbf{AFAT}~\cite{horn2019autofeat} overcomes the limitations of ERG by generating features multiple times and selecting them in multiple steps.
(4) \textbf{NFS}~\cite{chen2019neural} conceptualizes feature transformation as sequence generation and optimizes it using reinforcement learning.
(5) \textbf{TTG}~\cite{khurana2018feature} formulates the transformation process as a graph construction problem at the dataset level to identify optimal transformations.
(6) \textbf{GRFG}~\cite{xiao2024traceable} employs a cascading reinforcement learning structure to select features and operations, which ultimately generates new discriminative characteristics.
(7) \textbf{DIFER}~\cite{zhu2022difer} performs differentiable automated feature engineering in a continuous vector space, which is a gradient-based method.
(8) \textbf{FETCH}~\cite{li2023learning} is an RL-based end-to-end method that employs a single agent to observe the tabular state and make decisions sequentially based on its policy.
(9) \textbf{OpenFE}~\cite{zhang2023openfe} is an efficient method that initially evaluates the incremental performance of generated features and then prunes candidate features in a coarse-to-fine manner.
(10) \textbf{FastFT}~\cite{he2025fastft} incorporates the novelty into the reward function, accelerating the model’s exploration of effective transformations and improving the search productivity.

\subsection{Data Preparation}
To ensure experimental integrity, the datasets were divided into training and testing subsets to prevent data leakage and overfitting. 
The training dataset, comprising 80\% of the data, was used to optimize the reinforcement learning process. 
The testing datasets were hold-out used to evaluate the models' transformation and generation capabilities.
The partitioning principle was stratified sampling, which follows the same settings as in previous research~\cite{kdd2022,zhu2022difer}. 
Specifically, for regression tasks, we divided the labels into five ranges based on value size and randomly selected 20\% from each range for testing, with the remaining portion used for training. 
For classification tasks, we selected 20\% from each class for testing, with the remaining data used for exploration.
In the model's final evaluation phase, we applied the n-fold cross-validation method to partition the data and used the scikit-learn toolkit to test the hold dataset with the generated features for evaluation. 
The basic evaluation of downstream machine learning task scores was conducted utilizing the Random Forest Regressor and Classifier. 

\subsection{Hyperparameter Settings and Reproducibility}
\label{hyperparameter}
To comprehensively explore the feature space, we conducted exploration training for 50 episodes, each consisting of 100 steps, during the reinforcement learning agent optimization phase.
Following optimizing, we assessed the exploit ability of the collaborative agents by conducting 10 application episodes, each comprising 100 steps.
Following existing research~\cite{wang2024reinforcement,xiao2024traceable}, we set the number of clusters \( k \) to the square root of the current number of nodes, while the number of nodes triggered for pruning \( K \) is set to four times the original number of features. 
During step-wise pruning, we utilize the \( k \) most importance features.
We utilized a two-layer RGCN as the encoder for the transformation roadmap, and an embedding layer for the operation encoder.
The hidden state sizes for the roadmap encoder and operation encoder were set to 32 and 64, respectively.
Each agent was equipped with a two-layer feed-forward network for the predictor, with a hidden size of 100.
The target network was updated every 10 exploration steps by copying parameters from the prediction network.
To train the cascading agents, we set the memory buffer to 16 and the batch size to 8, with a learning rate of 0.01.
For the first 30\% epochs, we employed a node-wise pruning strategy to eliminate low-quality features.
Subsequently, we utilized a step-wise backtracking strategy for the remaining epochs to restore the optimal roadmap.
All experiments were conducted on the Ubuntu 18.04.6 LTS operating system, AMD EPYC 7742 CPU, and 8 NVIDIA A100 GPUs, with the framework of Python 3.8.18 and PyTorch 2.2.0~\cite{pytorch}.

\section{experiment}
\subsection{Overall Comparison}
This experiment aims to answer the question: \textit{Can our framework generate high-quality features to improve the downstream machine learning model?}
Table~\ref{main_table} presents the overall comparison between our model and other models in terms of F1-score for classification tasks and 1-RAE for regression tasks. 
We observed that our model outperforms other baseline methods in most datasets. 
The primary reason is that it dynamically captures and applies transformations across various stages of the feature transformation process rather than being restricted to the latest nodes, thereby enhancing flexibility and robustness. 
Compared to expansion-reduction such as OpenFE, our technique demonstrates an advantage in performance. 
The fundamental mechanism is that the reinforcement agent is capable of learning and refining its approach to the process, thereby achieving superior performance compared to random exploration. 
Another observation is that our model performs better than other iterative-feedback approaches, such as NFS, TTG, GRFG and FETCH.
An explanation could be that our model identifies and incorporates hidden correlations and mathematical properties, enabling it to develop an improved strategy for feature transformation, drawing on extensive historical knowledge from previous efforts. 
Compared with the AutoML-based approach DIFER, our technique demonstrates an improvement. 
This is primarily because DIFER relies on randomly generated transformations, which are unstable and prone to suboptimal results.
Overall, this experiment demonstrates that \model\ is effective and robust across diverse datasets, underscoring its broad applicability for automated feature transformation tasks.
\subsection{Significance of the Transformation Roadmap}
\label{graphablation}
\begin{figure*}[!htbp]
    \centering
    \vspace{-0.2cm} 
    \subfloat{\includegraphics[width=0.21\linewidth]{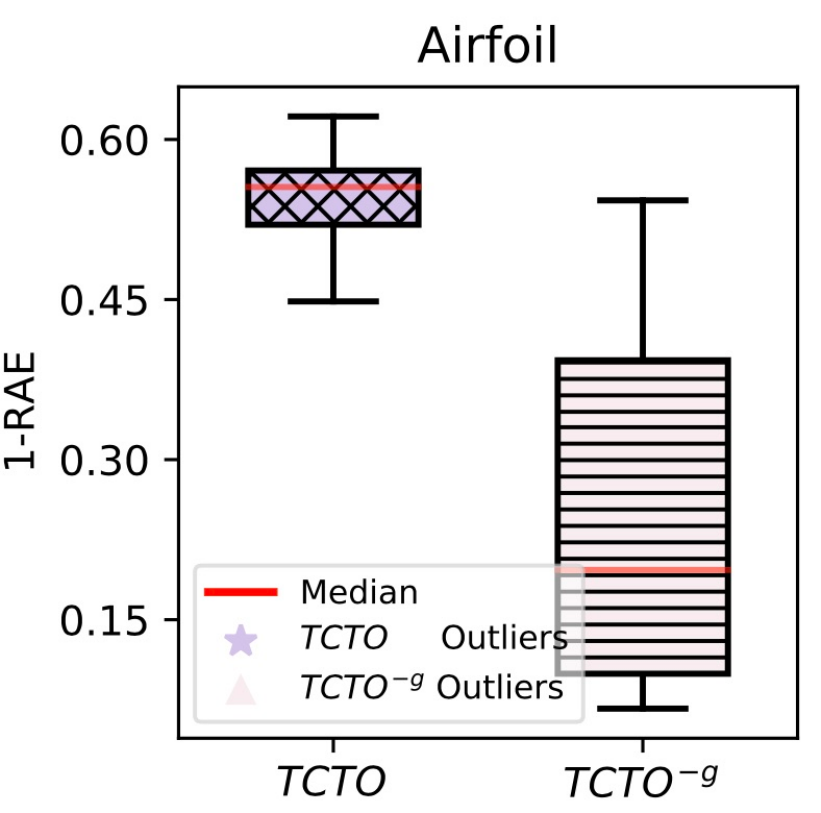}\label{fig:airfoil2}}
    \hfill
    \subfloat{\includegraphics[width=0.21\linewidth]{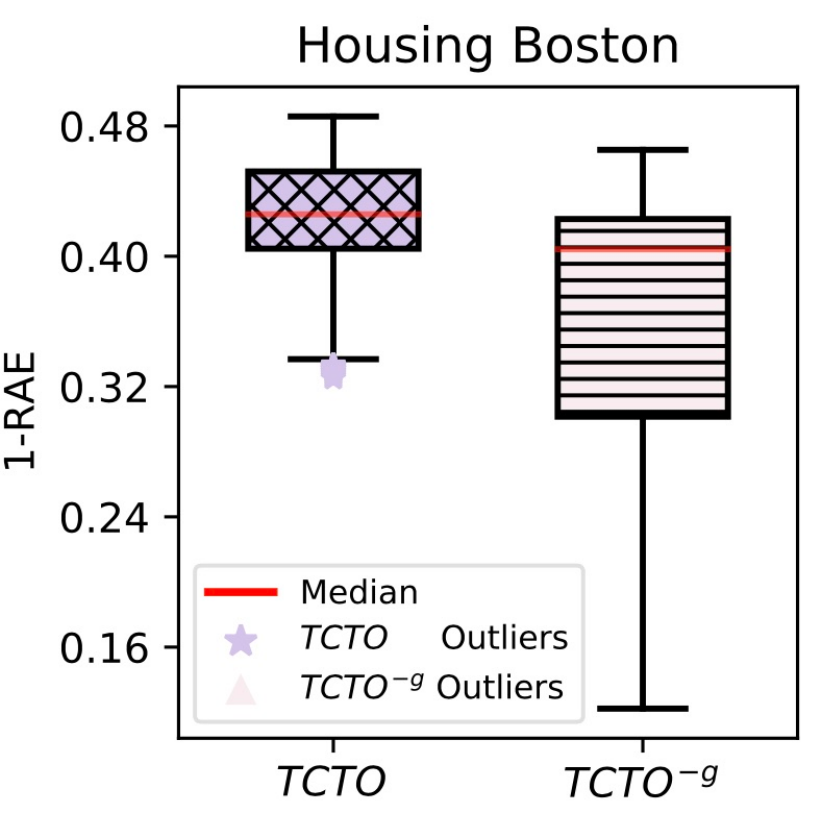}\label{fig:housing2}}
    \hfill
    \subfloat{\includegraphics[width=0.21\linewidth]{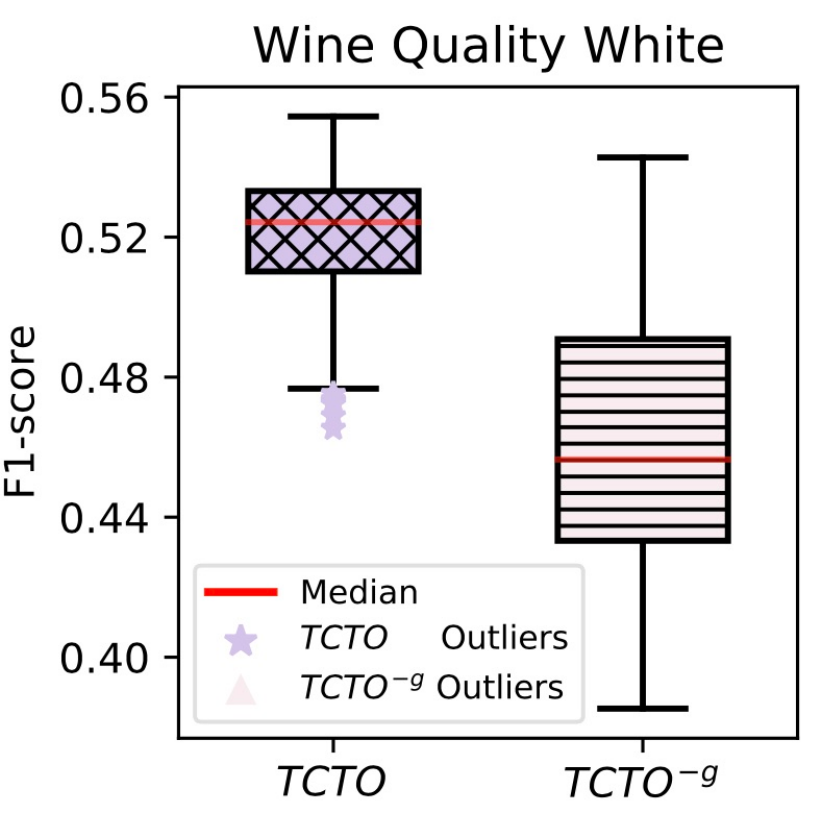}\label{fig:wqw2}}
    \hfill
    \subfloat{\includegraphics[width=0.21\linewidth]{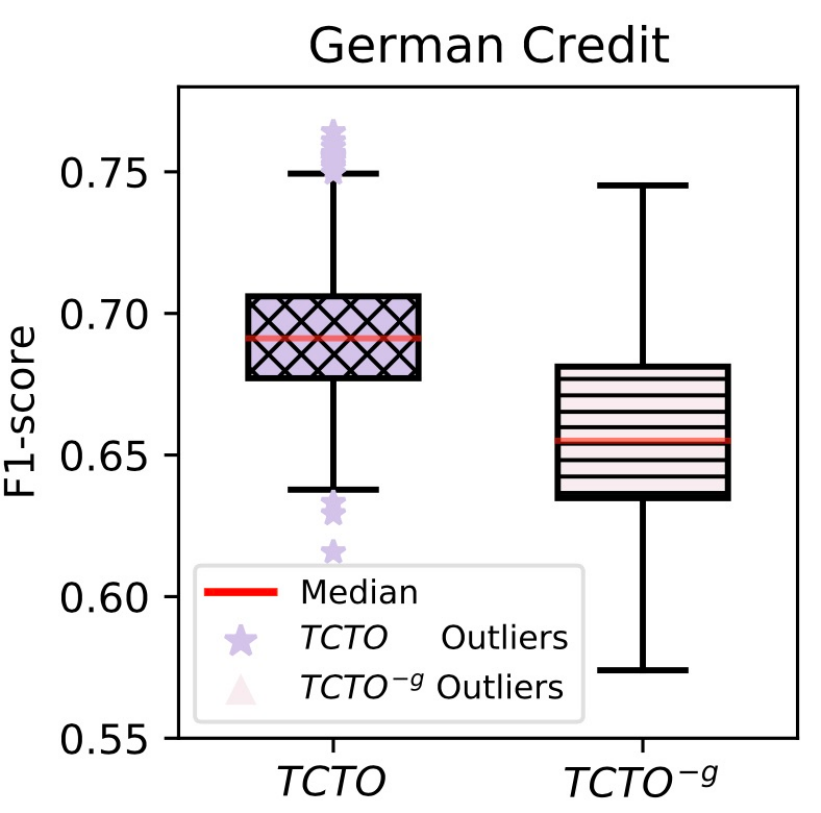}\label{fig:german2}}
    \caption{Stability comparison of TCTO and \(\text{model}^{-g}\) in four different datasets.}
    \label{exp:2}
    \vspace{-0.2 in} 
\end{figure*}

\begin{figure*}[!htbp]
    \centering
    \subfloat{\includegraphics[width=0.21\linewidth]{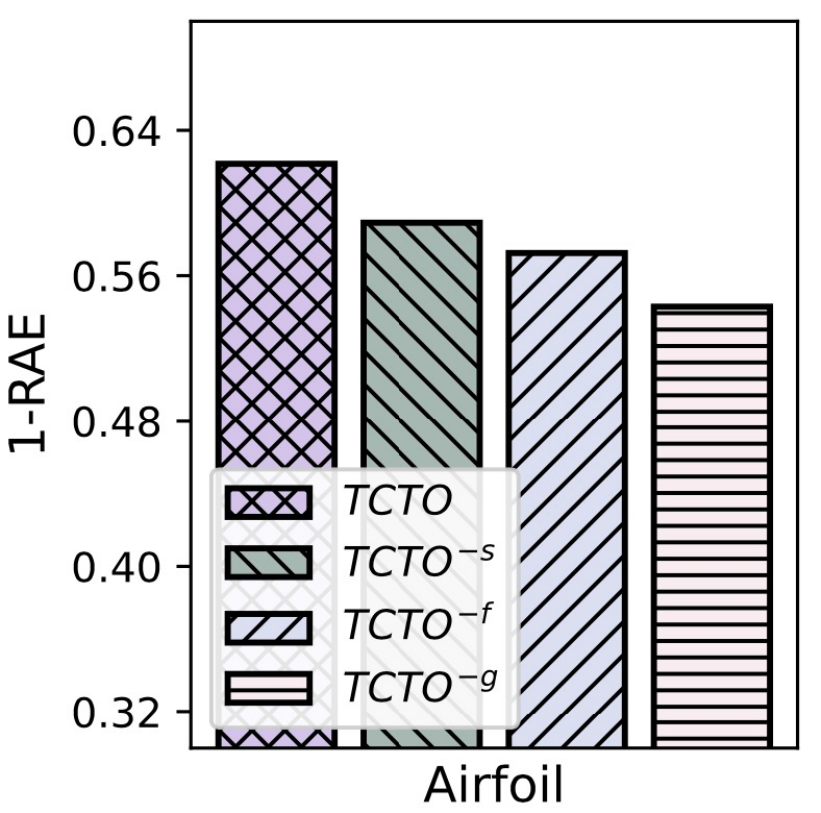}\label{fig:airfoil}}
    \hfill
    \subfloat{\includegraphics[width=0.21\linewidth]{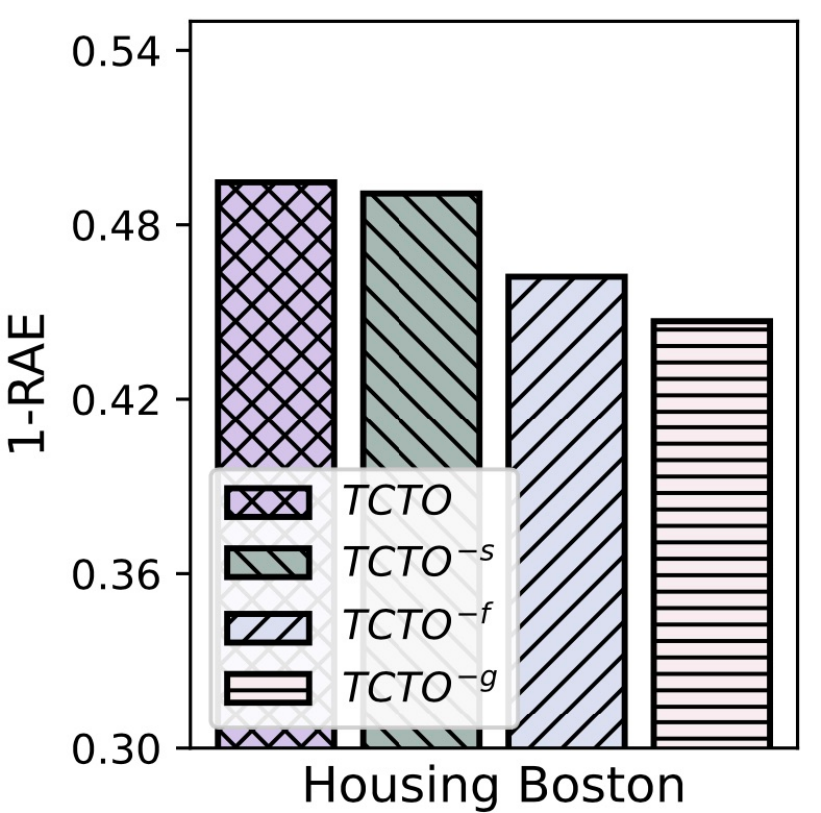}\label{fig:housing}}
    \hfill
    \subfloat{\includegraphics[width=0.21\linewidth]{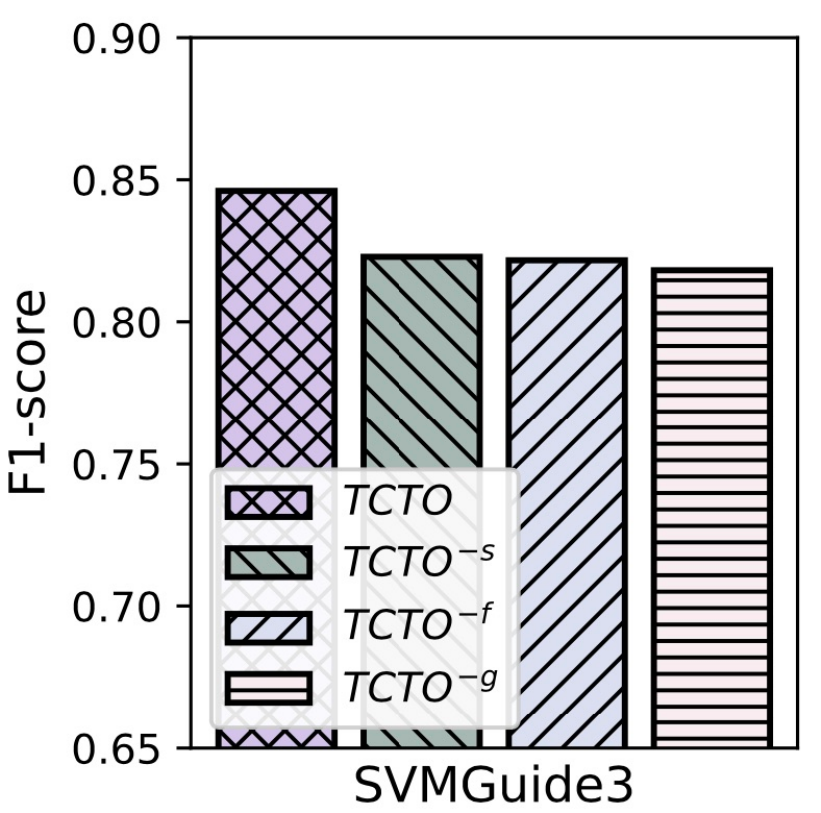}\label{fig:svmguide3}}
    \hfill
    \subfloat{\includegraphics[width=0.21\linewidth]{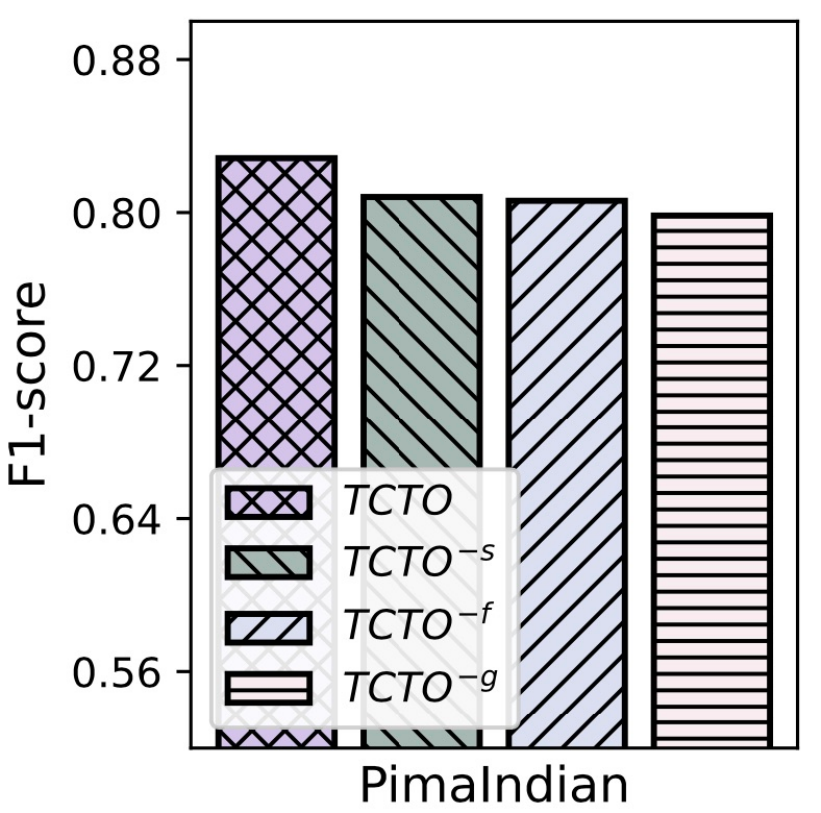}\label{fig:pima}}
    \caption{Comparison of TCTO and its variants in Regression and Classification tasks.}
    \label{exp:1}
    \vspace{-0.2in} 
\end{figure*}

This experiment aims to answer the question: \textit{How does the transformation roadmap impact each component in our model?}
We design three different ablation variants:
1) \textbf{\({\model}^{-f}\)} indicates that the clustering module ignores the mathematical \textbf{\underline{f}}eatures. 
2) \textbf{\({\model}^{-s}\)} indicates that the clustering module ignores \textbf{\underline{s}}tructural information. 
3) \textbf{\({\model}^{-g}\)} ablate the roadmap, remove the \textbf{\underline{g}}raph pruning, and directly adopt dataset statistic information (i.e., without RGCN component) as state representation.
The comparison results of these variants are reported in Figure~\ref{exp:2}-\ref{exp:1}. 

\noindent\textbf{Impact on Exploration Stability: }\label{stable}
We collected the performance of the downstream task at each exploration step of \model\ and the ablation variation method \({\model}^{-g}\).
Figure~\ref{exp:2} displays box plots summarizing the distributional characteristics of the experimental results.
We observe that the median line of our model is consistently higher than ${\model}^{-g}$.
Additionally, the interquartile range (IQR), depicted by the length of the box, indicates that our model's performance distribution is more concentrated than the ablation variation.
The observed stability in our model can be attributed to two primary factors. Firstly, the incorporation of historical and feature information within the roadmap provides guidance, steering the model towards more stable exploration directions. Secondly, the implementation of a roadmap pruning strategy alongside a backtracking mechanism plays a crucial role. It eliminates ineffective transformed features or reverts the model to the optimal state of the current episode, thereby ensuring stability throughout the exploration process.

\smallskip
\noindent\textbf{Impact on Exploration Effectiveness: }
Figure~\ref{exp:1} illustrates the effectiveness of the optimal features produced by our model and its variants in downstream tasks on the test dataset. 
We discovered that \model\ against the other three variants, while ${\model}^{-g}$ showed the weakest performance. 

\noindent\textit{Impact on Roadmap Modeling Approach:} 
From Figure~\ref{exp:1}, we found a decrease in the performance of downstream tasks when the roadmap structure is excluded, i.e., \({\model}^{-g}\). 
This performance decline is attributed to the loss of essential information that the transformation roadmap maintained. 
In contrast, utilizing the roadmap can enable agents to make strategic decisions based on comprehensive historical insights and complex feature interactions. 
This also indicates that the integration of roadmap structure and feature information is vital for a precise clustering, which can help the agents to organize transformation between two distinct groups of features, thus generating high-quality features. 

\noindent\textit{Impact on Clustering Component: } 
We can also observe that ${\model}^{-s}$ slightly outperforms ${\model}^{-f}$ on each task and dataset, i.e., the mathematical characteristic of the generated feature seems to be more significant than structural information. 
The underlying reason is that the mathematical characteristics of the generated features play a more crucial role in improving the clustering component compared to structural information from historical transformations.



\subsection{Scalability Analysis on Large-Scale Datasets}
\label{scalability}

This experiment aims to answer the question: \textit{Can our approach scale to large-scale datasets?} 
We categorize large-scale datasets into two types: large-sample and high-dimensional datasets.\\
\textbf{Scalability with large-sample datasets:} ALBERT is a large-sample dataset with 425,240 samples and 78 features. The time required for the downstream task is approximately 16 minutes per step which is unacceptable.
To address this issue, we switched to a more efficient downstream model LightGBM, which offers faster speed.
Table~\ref{main_table} demonstrates that \model\ can effectively scale by leveraging more efficient models for large-sample datasets.
We observed that all methods exhibit limited gains, and the results suggest that feature transformation methods have limited impact on extremely large-sample datasets, consistent with existing work (see Table 3 in study~\cite{zhang2023openfe}).
The key reason for this is that neural networks can learn latent patterns from sufficient samples, making additional feature transformation less essential.

\noindent\textbf{Scalability with high-dimensional datasets:} Newsgroups is a high-dimensional dataset with 13,142 samples and 61,188 features. The time required for the downstream task is approximately 5 minutes per step.
For high-dimensional datasets, we employed a pruning strategy to remove irrelevant root nodes before exploring the dataset.
Table~\ref{main_table} shows that \model\ outperforms baseline methods in terms of Macro-F1. The pruning strategy helps mitigate the complexity of high-dimensional datasets, speeding up the process while maintaining performance.\\
In conclusion, experiments demonstrate that \model\ is scalable and performs well on both large-sample and high-dimensional datasets when appropriate strategies are employed.

\begin{figure}[!t]
\centering
\centerline{\includegraphics[width=0.7\columnwidth]{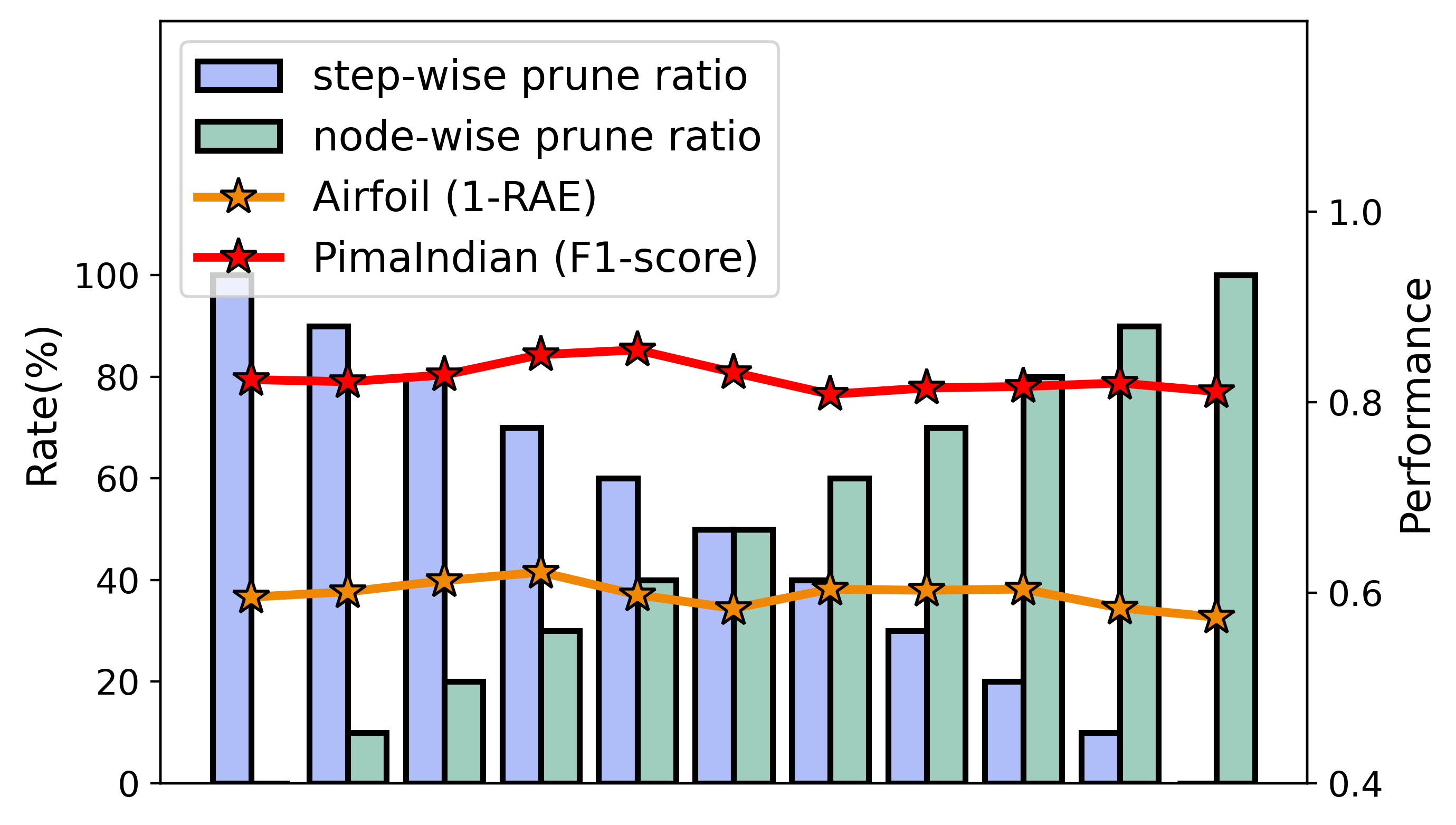}}
\caption{Study of the node-wise and step-wise pruning ratio on Airfoil and PimaIndia datasets.}
\label{prune_param}
\vspace{-0.5cm}
\end{figure}

\subsection{Analysis on Graph Pruning Technique}\label{graph_pru}

This experiment aims to answer the question: \textit{How to balance the trade-off between node-wise and step-wise pruning ratios?} 
To validate the pruning ratio sensitivity of our model, we set the ratio from 0 to 1 to observe the differences.  
We report the performance variations on Airfoil (regression task) and PimaIndia (classification task) in Figure~\ref{prune_param}. 
We observed that adopting more node-wise pruning, downstream ML performance improves initially and then declines. 
A possible reason is that the node-wise pruning could preserve search space diversity when agents are unfamiliar with the dataset. 
However, with more application of the node-wise pruning strategy, \model{} cannot backtrack to the previous optimal transformation roadmap, resulting in suboptimal paths and decreased performance. 
We set the node-wise pruning ratio to 30\% according to the experimental results.

\begin{figure}[!t]
    \centering
    \subfloat{
        \includegraphics[width=0.75\linewidth]{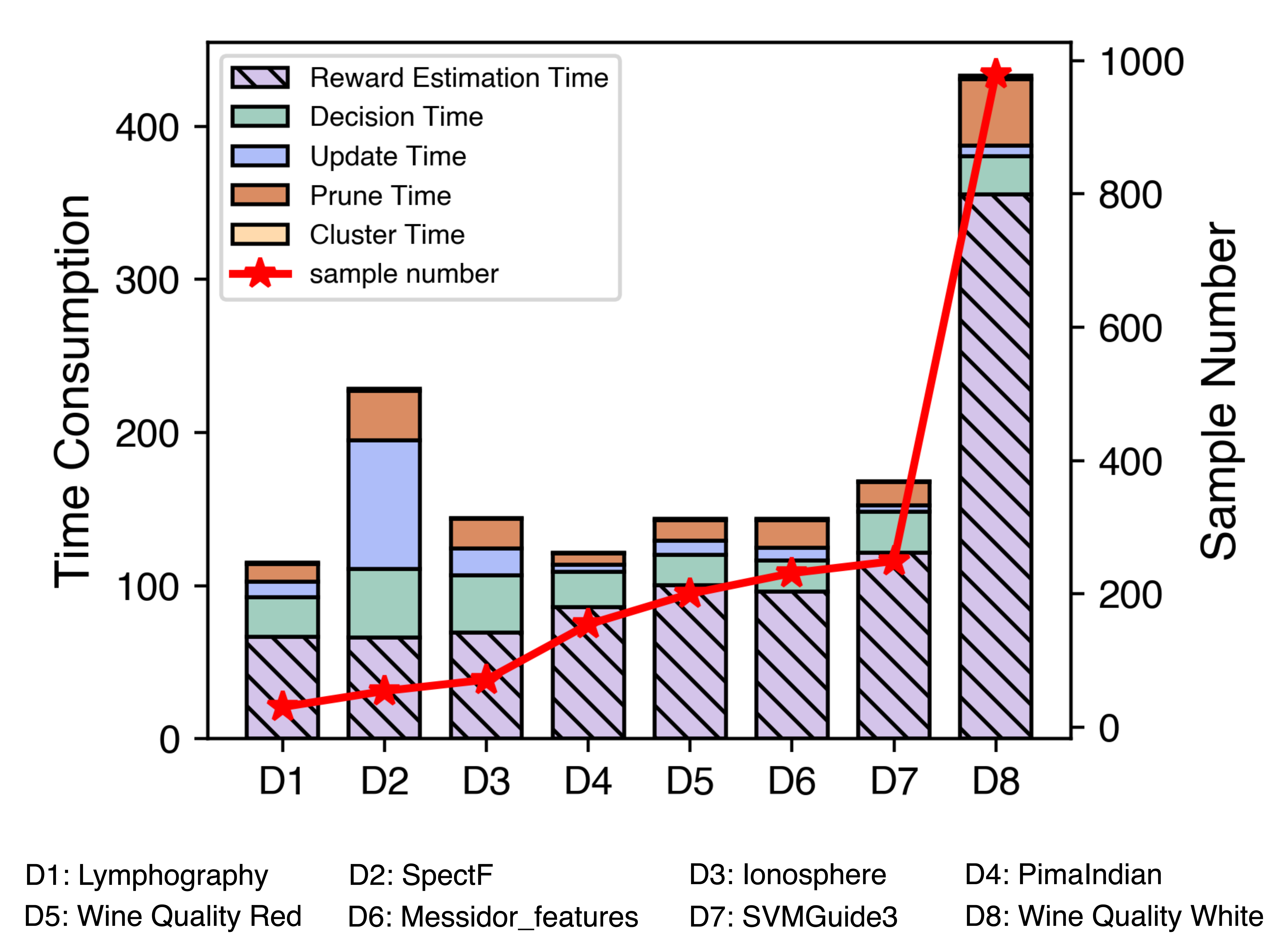}
    }
    \vfill 
    \subfloat{
        \includegraphics[width=0.75\linewidth]{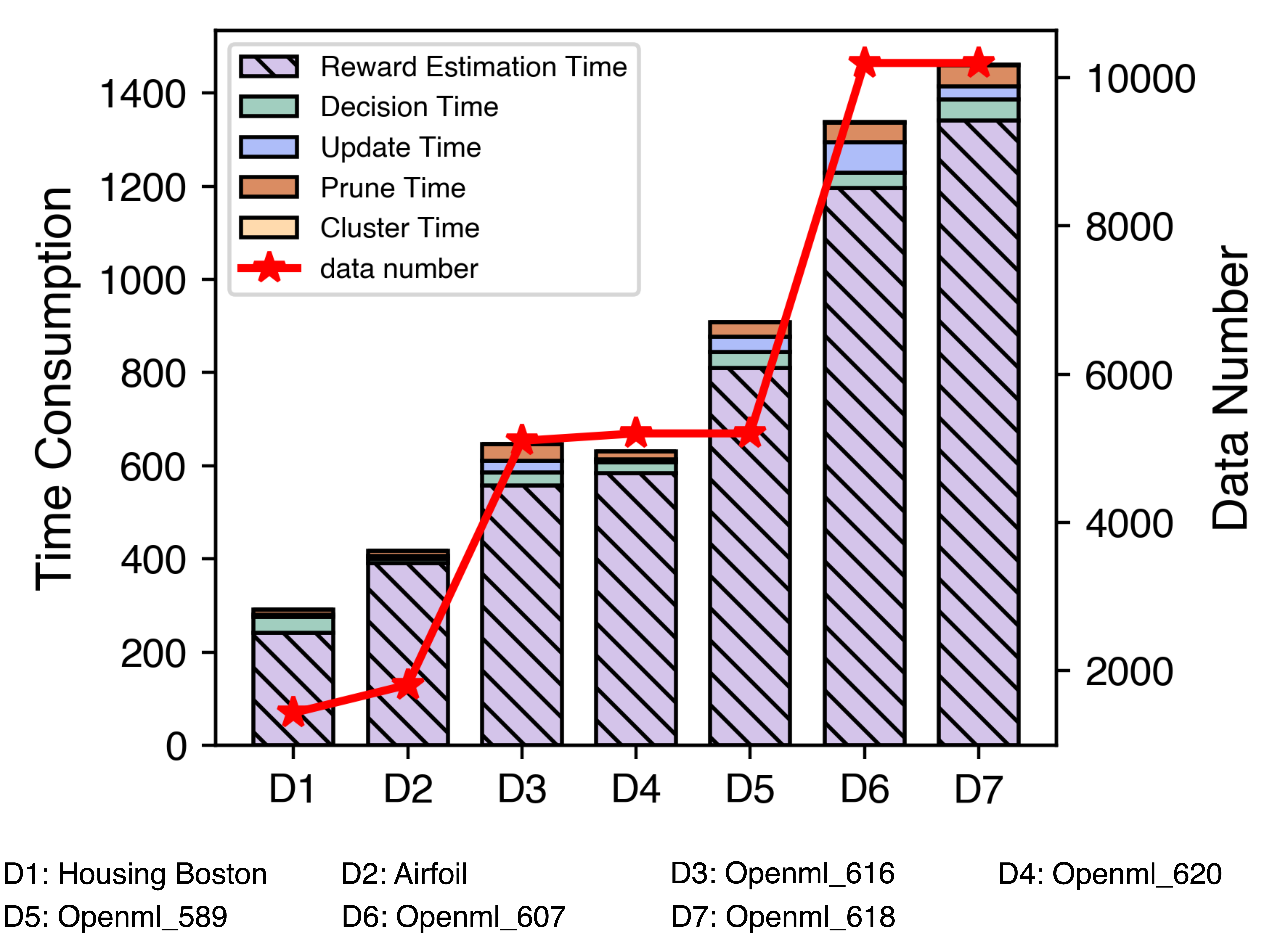}
    }
    \caption{Time consumption of \text{model} on different tasks.}
\label{fig:time_consumption}
\end{figure}
\subsection{Runtime Complexity and Bottleneck Analysis}
\label{runtime}
This experiment aims to answer: \textit{What is the main temporal bottleneck of \model?}
Figure~\ref{fig:time_consumption} visualized the average empirical running time consumption on each dataset of different modules to analyze the runtime efficiency, including reward estimation, agent decision-making, roadmap updating, clustering and pruning. 
We observed that the reward estimation time dominates the overall time consumption across all dataset sizes. 
This phenomenon can be primarily attributed to the computationally intensive nature of the downstream tasks evaluation process. 
In addition, the time cost of reward estimation increases proportionally with the size of the dataset, resulting in a linear scalability of \model\ in terms of time complexity. 
In summary, the main temporal bottleneck of this framework, as well as other iterative-feedback approaches, is the downstream task evaluation in the reward estimation component.

\subsection{Analysis on the weight of reward function}\label{reward_weight}

\begin{figure}[!t]
\centering
\centerline{\includegraphics[width=0.85\columnwidth]{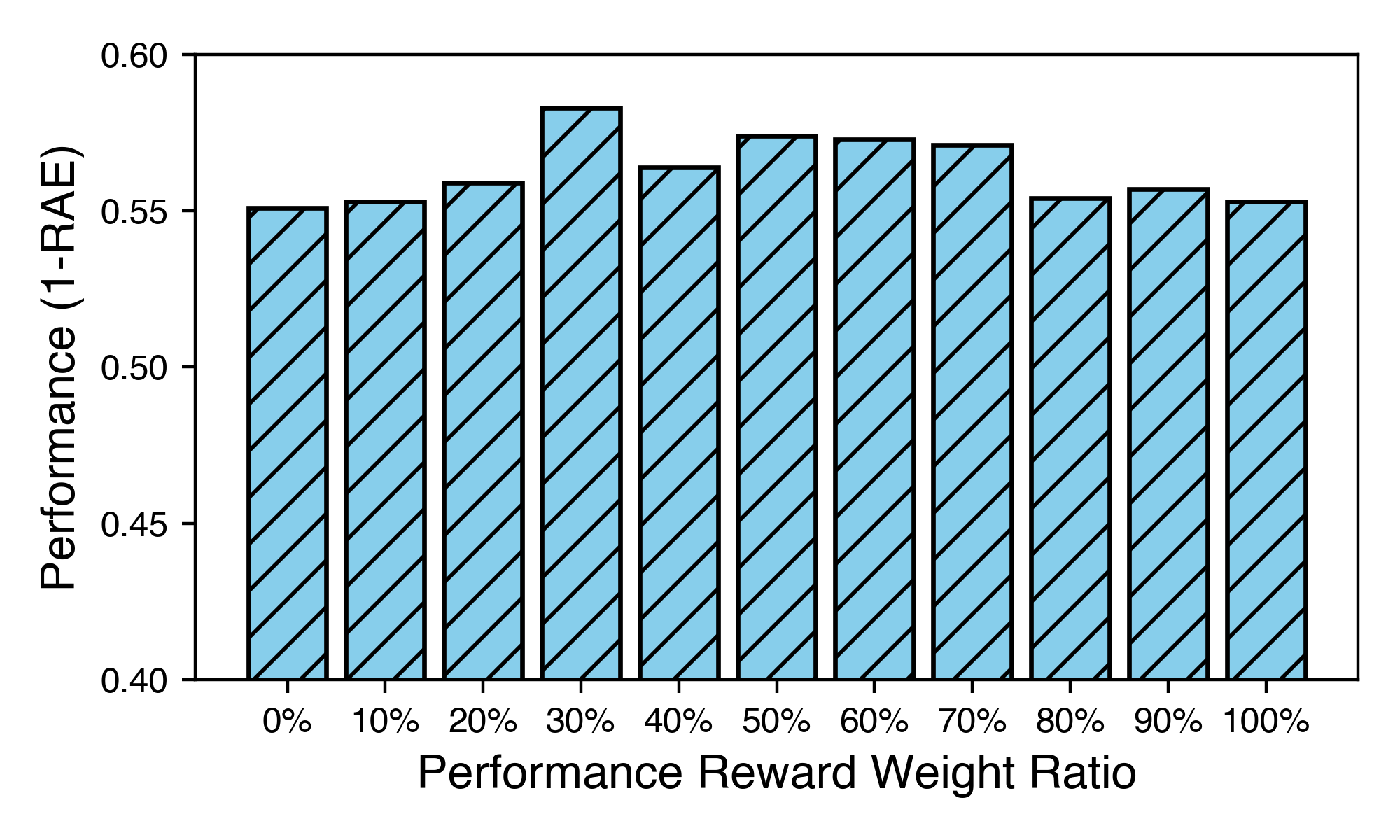}}
\caption{Impact of weights between performance and complexity rewards.} 
  \label{reward_ratio}
\end{figure}
This experiment aims to answer the question: \textit{How does the trade-off between performance and complexity impact the performance?}
We conducted a preliminary experiment using the Airfoil dataset to analyze the impact of reward weights during the optimization stage.
Figure~\ref{reward_ratio} shows that only the complexity or the performance reward is used exclusively, the performance is noticeably lower. 
This result suggests that while performance rewards encourage agents to generate high-value features, overly complex features can be detrimental to the downstream task. 
With a balanced weight of them, the performance fluctuates slightly. 
Based on these preliminary results, we concluded that a ratio of 1:1 between feature quality and complexity, providing stable and reliable performance.

\label{robust_check}
\begin{table}[!h]
\centering
\caption{Robustness check with distinct ML models on Housing Boston (HB) and Messidor\_features (MF) datasets}
\label{robust}
\begin{minipage}[t]{\linewidth}
\centering
\resizebox{\linewidth}{!}{
\begin{tabular}{cccccccc}
\toprule 
 HB     & RFR   & Lasso & XGBR  & SVM-R & Ridge-R & DT-R  & MLP   \\ \midrule
AFAT   & 0.416 & 0.277 & 0.347 & 0.276 & 0.187   & 0.161 & 0.197 \\ 
ERG   & 0.409 & 0.162 & 0.331 & 0.278 & 0.256   & 0.257 & 0.300 \\ 
NFS   & 0.425 & 0.169 & 0.391 & 0.324 & 0.261   & 0.293 & 0.306 \\ 
RDG   & 0.404 & 0.193 & 0.299 & 0.287 & 0.218   & 0.257 & 0.279 \\ 
TTG   & 0.396 & 0.163 & 0.370 & 0.329 & 0.261   & 0.294 & 0.308 \\ 
GRFG  & 0.465 & 0.185 & 0.435 & 0.363 & 0.265   & 0.197 & 0.208 \\ 
FastFT & 0.491 & 0.238 & 0.394 & 0.332 & 0.297 & 0.292 & \textbf{0.310} \\
\textbf{TCTO} & \textbf{0.495} & \textbf{0.370} & \textbf{0.444} & \textbf{0.384} & \textbf{0.317} & \textbf{0.350} & \textbf{0.310} \\ \bottomrule
\end{tabular}}
\end{minipage}\vfill
\begin{minipage}[t]{\linewidth}
\centering
\medskip
\resizebox{\linewidth}{!}{
\begin{tabular}{cccccccc}
\toprule
 MF    & RFC   & XGBC  & LR    & SVM-C & Ridge-C & DT-C  & KNB   \\ \midrule
AFAT  & 0.665 & 0.608 & 0.634 & 0.664 & 0.633   & 0.564 & 0.530 \\
ERG  & 0.669 & 0.703 & 0.659 & 0.571 & 0.654   & 0.580 & 0.537 \\
NFS  & 0.638 & 0.607 & 0.627 & 0.676 & 0.646   & 0.613 & 0.577 \\
RDG  & 0.624 & 0.607 & 0.623 & 0.669 & 0.660   & 0.609 & 0.577 \\
TTG  & 0.655 & 0.607 & 0.633 & 0.676 & 0.646   & 0.599 & 0.577 \\
GRFG & 0.718 & 0.648 & 0.642 & 0.486 & 0.663   & 0.580 & 0.552 \\
FastFT & 0.731 & 0.723 & 0.694 & 0.681 & 0.678 &  0.649 & \textbf{0.587} \\
\textbf{TCTO} & \textbf{0.742} & \textbf{0.730} & \textbf{0.706} & \textbf{0.701} & \textbf{0.689} & \textbf{0.652} & \textbf{0.587} \\ \bottomrule
\end{tabular}}

\end{minipage}
\vspace{-0.5cm}
\end{table}

\subsection{Robustness Check}
This experiment aims to answer the question: \textit{Are our generative features robust across different machine learning models used in downstream tasks?}
We evaluate the robustness of the generated features on several downstream models.
For regression tasks, we substitute the Random Forest Regressor (RFR) with Lasso, XGBoost Regressor (XGB), SVM Regressor (SVM-R), Ridge Regressor (Ridge-R), Decision Tree Regressor (DT-R), and Multilayer Perceptron (MLP).
For classification tasks, we assess the robustness using Random Forest Classifier (RFC), XGBoost Classifier (XGBC), Logistic Regression (LR), SVM Classifier (SVM-C), Ridge Classifier (Ridge-C), Decision Tree Classifier (DT-C), and K-Neighbors Classifier (KNB).
Table~\ref{robust} presents the results in terms of 1-RAE for the Housing Boston dataset and F1-score for the Messidor\_features dataset, respectively.
We can observe that the transformed features generated by our model consistently achieved the highest performance in regression and classification tasks among each downstream machine learning method.
The underlying reason is that these features contain significant information that is capable of fitting into various machine learning tasks.
Therefore, this experiment validates the effectiveness of our model in generating informative and robust features for various downstream models.

\section{Case Study on Generated Features}
\label{case_study}

\begin{table*}[t]
\centering
\caption{the ten most significant features of original and transformed datasets for Housing Boston and White Wine Quality}
\vskip -0.2in
\label{trace}
\begin{center}
\begin{small}
\begin{minipage}[t]{0.7\linewidth}
\centering
\medskip
\resizebox{\linewidth}{!}{
\begin{tabular}{rcrccc}
\toprule
\multicolumn{2}{c}{Housing Boston}             & \multicolumn{2}{c}{TCTO$^{-g}$}               & \multicolumn{2}{c}{TCTO}                                   \\
\multicolumn{1}{c}{feature} & importance & \multicolumn{1}{c}{feature} & importance & feature                                       & importance \\ \hline
lstat                       & 0.362      & quan\_trans(lstat)          & 0.144      & \cellcolor[HTML]{C0C0C0}$v_{18}:\sqrt{|v_{17}|}$       & 0.080     \\
rm                          & 0.276      & lstat                       & 0.135      & \cellcolor[HTML]{C0C0C0}$sta(v_{17})$ & 0.077      \\
dis                         & 0.167      & quan\_trans(rm)             & 0.126      & \cellcolor[HTML]{C0C0C0}$sta(\sqrt{|v_{17}}|)$                          & 0.054      \\
crim                        & 0.072      & rm                          & 0.119      & \cellcolor[HTML]{C0C0C0}$sta(v_{16})$                                         & 0.054      \\
rad                         & 0.032      & (dis+(...))-quan(lstat)     & 0.076      & \cellcolor[HTML]{C0C0C0}$sta(\sqrt{\sqrt{v_{18}}})$                   & 0.053      \\
black                       & 0.032      & (dis*(...))+(...)+(dis+...) & 0.050      & \cellcolor[HTML]{C0C0C0}$v_{16}:\frac{1}{\sin{v_{12}}-v_{0}}$                                   & 0.053      \\
age                             & 0.030       & (dis+...)+(...)-(zn+(...))      & 0.048       & $sta(v_{24})$ & 0.050       \\
nox                         & 0.011      & (dis+...)-(...)+quan(rm)    & 0.028      & $\min(v_5)$                                   & 0.044      \\
ptratio                     & 0.007      & (dis+..lstat)-(...+rad)     & 0.016      & \cellcolor[HTML]{C0C0C0}$v_{17}:\sqrt{|v_{16}|}$ & 0.037      \\
indus                       & 0.005      & (dis+..crim)-(...+rad)      & 0.015      & $v_{12}$                    & 0.025      \\ \midrule
\multicolumn{1}{c}{1-RAE:0.414} & Sum:0.993 & \multicolumn{1}{c}{1-RAE:0.474} & Sum:0.757 & 1-RAE:0.494                                                 & Sum:0.527 \\ \bottomrule
\end{tabular}}
\end{minipage}\hfill
\begin{minipage}[t]{0.7\linewidth}
\centering
\medskip
\resizebox{\linewidth}{!}{
\begin{tabular}{rcrccc}
\toprule
\multicolumn{2}{c}{Wine Quality White}             & \multicolumn{2}{c}{TCTO$^{-g}$}                   & \multicolumn{2}{c}{TCTO}                                        \\
\multicolumn{1}{c}{feature} & importance & \multicolumn{1}{c}{feature}     & importance & feature                                            & importance \\ \hline
alcohol                     & 0.118      & quan\_trans(alcohol)            & 0.043      & \cellcolor[HTML]{C0C0C0}$v_{2}+v_{30}$             & 0.026      \\
density                     & 0.104      & alcohol                         & 0.036      & \cellcolor[HTML]{C0C0C0}$\sin{(\sin{(v_{0})})}+v_{30}$ & 0.025      \\
volatile            & 0.099      & ((den...)+(alc...)/(...))       & 0.028      & \cellcolor[HTML]{C0C0C0}$v_{5}+v_{30}$             & 0.024      \\
free sulfur                & 0.093       & quan\_trans(density)               & 0.028       & \cellcolor[HTML]{C0C0C0}$\sin{(v_{0})}+v_{30}$ & 0.023       \\
total sulfur       & 0.092      & density                         & 0.028      & $v_{2}$                                            & 0.023      \\
chlorides                   & 0.091      & (den/(...))+(dens...)/(...)     & 0.026      & \cellcolor[HTML]{C0C0C0}$v_{3}+v_{30}$             & 0.023      \\
residual              & 0.087      & (den/(...)+((...)/tan(...))     & 0.024      & \cellcolor[HTML]{C0C0C0}$v_{6}+v_{30}$             & 0.021      \\
pH                          & 0.082      & (den/...)-(...+stand(...))      & 0.023      & \cellcolor[HTML]{C0C0C0}$v_{7}+v_{30}$             & 0.021      \\
citric acid                 & 0.081      & (citr/(...)+(...)/(tanh(...)) & 0.023      & \cellcolor[HTML]{C0C0C0}$v_{0}+v_{30}$             & 0.021      \\
fixed acidity       & 0.078      & (free/(...)+(...)/tanh(...))   & 0.023      & \cellcolor[HTML]{C0C0C0}$v_{11}+v_{30}$            & 0.021      \\ \midrule
\multicolumn{1}{c}{F1-score:0.536} & Sum:0.924 & \multicolumn{1}{c}{F1-score:0.543} & Sum:0.282 & F1-score:0.559                               & Sum:0.228 \\ \bottomrule
\end{tabular}}
\end{minipage}
\end{small}
\end{center}
\vskip -0.2in
\end{table*}

\noindent This experiment aims to answer the question: \textit{Can our model reuse high-value sub-transformations and generate high-quality feature space?}
Table~\ref{trace} presents the Top-10 most important features generated from the original dataset, our proposed method, and its feature-centric variant (\model$^{-g}$). 
We first observe that \model\ consistently reuses many high-value sub-transformations (\textit{highlighted in \colorbox{gray!30}{gray} blocks}), such as node $v_{17}$ in Housing Boston and node $v_{30}$ in Wine Quality White.
The repeated appearance of these features not only indicates their strong contribution to ML tasks, but also demonstrates our model’s ability to effectively identify and reuse important intermediate features. 
Compared to \model$^{-g}$, our model tends to transform and reuse meaningful intermediate nodes more frequently, leading to the generation of more significant features. 
By leveraging historical transformation paths in the roadmap, the model identifies optimal substructures and combines them.
Additionally, the importance scores of the transformed features in our model are more evenly distributed than those in the original dataset and its variant.
Since our model has better performance, we speculate that our framework comprehends the properties of the feature set and ML models to produce numerous significant features by combining the original features. 
Finally, the explicit transformation records presented in Table~\ref{trace}, formulated by combinations of both original and intermediate features, ensure full traceability.
Such characteristics of traceability may assist domain experts in uncovering novel mechanisms in their respective fields.

\section{Related work}
Feature transformation refers to the process of handling and transforming raw features to better suit the needs of machine learning algorithms \cite{hancock2020survey,chen2021techniques}. 
Automated feature transformation implies that machines autonomously perform this task without the need for human prior knowledge\cite{lam2017one}.
There are three mainstream approaches:
The expansion-reduction based method~\cite{kanter2015deep,horn2019autofeat,khurana2016cognito,lam2017one,khurana2016automating}, characterized by its greedy or random expansion of the feature space\cite{katz2016explorekit,dor2012strengthening}, presents challenges in generating intricate features, consequently leading to a restricted feature space. 
The iterative-feedback approach~\cite{khurana2018feature,tran2016genetic,kdd2022,xiao2022traceable,zhu2022evolutionary,xiao2024traceable} methods integrate feature generation and selection stages into one stage learning process, and aims to learn transformation strategy through evolutionary or reinforcement learning algorithms~\cite{ren2023mafsids}.
However, these methods usually model the feature generation task as a sequence generation problem, ignoring historical and interactive information during the transformation progress, result in lack of stability and flexibility.
The AutoML-based approaches~\cite{wang2021autods,zhu2022difer,xiao2023discrete,ying2023self} have recently achieved significant advancement.
However, they are limited by the quality of the collected transformation and also the lack of stability and traceability during the generation phase. 
To overcome these problems, \model\ introduces a novel framework that integrates structural insights based on roadmaps and a backtracking mechanism with deep reinforcement learning techniques to improve feature engineering.

\section{discussion}
We outline two future works: incorporating large language models and exploring further application scenarios.

\subsection{Large Language Models as Data Scientists}
Recently some research has utilized Large Language Model(LLM) to generate high-quality features~\cite{hassan2023chatgpt, long2024llms, zhang2024dynamic, hollmann2024large}. 
However, LLMs exhibit limitations from two perspectives: the semantic understanding of feature names and the issue of hallucination.

\textbf{Regarding the Limitation of Feature Names Comprehension}:
LLM-based methods utilize the comprehensive ability, conducting transformation to generate high-quality features based on the semantic of feature names. 
However, due to data quality issues, feature names are frequently anonymous or missing in some case, such as federated learning scenarios.
With semantic blinding, ~\cite{hollmann2024large} found \textit{a strong drop in performance} based on their experimental conclusion.

\textbf{Regarding the Issue of Hallucination}:
When LLMs are engaged in certain feature transformation tasks, such as evaluating feature importance, they often encounter the hallucination problem. 
The hallucination problem in LLMs can result in \textit{the generation of irrelevant or unsupported content}~\cite{hassan2023chatgpt}.
Due to hallucination problems, ~\cite{jeong2024llmselectfeatureselectionlarge} reported that without providing datasets, LLMs can produce a precise numerical value for importance.

In contrast to these approaches, \model{} operates independently of the semantic context of feature names, allowing for broader application scenarios.
We also acknowledge that integrating common or injected knowledge in LLMs has the potential to greatly improve the scalability, performance, and efficiency of our framework. 
It also worth noting that  ~\cite{jeong2024llmselectfeatureselectionlarge}, GPT-4~\cite{openai2024gpt4technicalreport} exhibits notable hallucination problems. 
However, we have noticed a significant improvement following the release of GPT-4o.
In future work, we will explore how LLMs can be utilized to comprehend and evaluate the roadmap that identified by TCTO, focusing on research aspects such as node clustering,  reward feedback and node pruning.

\subsection{Application Scenarios of Feature Transformation}
For years, feature engineering has been a significant step~\cite{sambasivan2021everyone,strickland2022andrew} before applying computational statistics or machine learning methods to data, and its importance has grown with the growing trend of interdisciplinary research between artificial intelligence and various fields, such as life sciences~\cite{ofer2015profet,chen2020ilearn,chicco2022eleven,bonidia2022bioautoml} and material sciences~\cite{dai2020using,kaundinya2021machine,xiang2021physics,anand2022topological}. 
Given the complexity inherent in scientific data, automated feature transformation methods like the one proposed in this work have significant potential to advance various AI4Science disciplines.
In bio-informatics and computational biology, our approach can aid in extracting pivotal gene, protein, or metabolites combinations (as depicted in our motivation, Figure~\ref{fig:motivation}) from high-throughput sequencing data, enhancing the identification of gene networks associated with diseases. 
In the realm of chemistry and drug discovery, chemically meaningful features can automatically be generated to improve the accuracy of molecular activity and toxicity predictions, thereby accelerating the development of new pharmaceuticals. 
In future research, we plan to advance our study by assisting life sciences experts in identifying combinations within population cohort data.
The feature transformation roadmap can ensure traceability and interpretability, thereby aiding scientific discovery and enhancing the model's accuracy in early disease detection.

\section{conclusion}
We introduce TCTO, an automated framework for conducting feature transformation. 
Our approach focuses on managing transformation process through a roadmap, which tracks and organizes the process.
There are three main benefits to our approach:
(1) Enhanced Agility through Traceable Roadmap: Our approach maintains a traceable roadmap of all transformation steps. It can apply transformations across various stages by utilizing roadmap that contains structural and mathematical characteristics.
(2) Insightful Exploration through Historical Experience: The roadmap structure automatically logs all feature transformations, enabling efficient exploration by leveraging historical transformations, and ensuring a more insightful exploration process.
(3) Increased Robustness through Backtracking: The roadmap's built-in backtracking mechanism allows the framework to correct or change its path if it encounters inefficient or suboptimal transformations, thereby improving the model's robustness.
Extensive experiments show that \model\ is effective and flexible in feature transformation.

\balance
\normalem

\bibliography{reference}
\bibliographystyle{IEEEtran}
\end{document}